\documentclass[letterpaper, 10 pt, journal, twoside]{IEEEtran}
\usepackage{amsmath,amsfonts}
\usepackage{algorithmic}
\usepackage{algorithm}
\usepackage{array}
\usepackage{textcomp}
\usepackage{stfloats}
\usepackage{url}
\usepackage{verbatim}
\usepackage{graphicx}
\usepackage{cite}
\usepackage{caption}
\usepackage{multirow}
\usepackage{colortbl} 
\usepackage{xcolor}
\usepackage{subcaption}
\usepackage{arydshln}
\usepackage{placeins}
\usepackage{tabularray}
\usepackage{orcidlink}
\UseTblrLibrary{booktabs}
\usepackage{caption}
\usepackage{amsthm}
\usepackage{hyperref}
\theoremstyle{remark} 
\newtheorem{remark}{Remark} 

\begin{document}

\title{Towards~Adaptive~Interaction~Strategies~for~Human Companion~Robot~via~Deep~Reinforcement~Learning}

\author{Cong-Thanh Vu,~\IEEEmembership{Student Member,~IEEE} and Yen-Chen Liu,~\IEEEmembership{Senior Member,~IEEE}
\thanks{This work was supported in part by the National Science and Technology Council (NSTC), Taiwan, under Grant NSTC 114-2628-E-006-010 and NSTC 114-2218-E-006-021.}
\thanks{Cong-Thanh Vu and Yen-Chen Liu are with the Department of Mechanical Engineering, National Cheng Kung University, Tainan 70101, Taiwan. Email: {\href{mailto:vuthanh.cdt@gmail.com}{\texttt{vuthanh.cdt@gmail.com}}, \href{mailto:yliu@mail.ncku.edu.tw}{\texttt{yliu@mail.ncku.edu.tw.}}}}
}

\markboth{IEEE TRANSACTIONS ON SYSTEMS, MAN, AND CYBERNETICS: SYSTEMS}%
{VU \MakeLowercase{et al.}: Towards~Adaptive~Interaction~Strategies~for~Human Companion~Robot~via~Deep~Reinforcement~Learning}


\maketitle

\begin{abstract}
In the field of Human-Robot Interaction (HRI), achieving flexibility in human-accompanying within real-world environments holds great potential for various applications but also poses significant challenges. Traditional methods typically restrict robots to fixed positions relative to humans, such as tracking from behind, in front, or side-by-side, which limits robot adaptability in dynamic workspaces. This study introduces a novel human-companioning strategy that uses Reinforcement Learning (DRL) to enable mobile robots to dynamically adjust their tracking positions according to varying conditions. An interaction space is defined to capture the relationship between the human and the robot while considering the environment, which serves as the basis for state spaces in DRL to assist the robot in adapting to environmental changes. A human-robot companion controller is developed by integrating Model Predictive Path Integral (MPPI) control with Control Barrier Functions (CBF), ensuring that the robot accurately follows the target's movement in both position and orientation while avoiding obstacles and enhancing social acceptance and safety. The proposed approach is evaluated in real-world scenarios, both indoors and outdoors, and compared with other studies. The results show that the proposed method improves the success rate and tracking accuracy by at least 24\% and 47\%, respectively, while enhancing human comfort. Experiments demonstrate the robot’s ability to flexibly accompany a person walking at speeds of up to 1.7 m/s, dynamically adjusting its strategy without being confined to a fixed position. Additionally, the robot respects the human’s intimate space to ensure safety, comfort, and effective obstacle avoidance.
\end{abstract}

\section{Introduction}
\IEEEPARstart{N}{owadays}, in light of the increasing emphasis on mobile robots to enhance and support human activities, there has been a substantial drive to develop robots that can operate effectively across a diverse range of environments, from assisting with everyday tasks to providing logistical support \cite{10829984,11223901,9113661}, as well as serving as everyday service robots \cite{9984687}. As these robots are expected to perform in human-centric settings, advancing Human-Robot Interaction (HRI) has emerged as a key research priority \cite{10551849,11150767}. A critical enabler of this vision is the ability of robots to accurately accompany and interact with people \cite{li2023exploring}. Moreover, mobile robots with robust human-following capabilities have proven their utility in numerous sectors, including manufacturing, logistics, healthcare, and public spaces \cite{7637024,8492362,islam2019person}.

Human-following from behind is a widely adopted strategy in companion robotics. Chen et al. \cite{chen2019human} integrated gesture recognition with a PID controller to maintain the desired relative pose, while Wang et al. \cite{7852457} employed gait recognition for robust human tracking. Gui et al. \cite{10606944} proposed Deep-MPC to predict human motion and improve tracking performance. To further enhance navigation in dynamic environments with obstacle avoidance, Toan et al. \cite{van2023human} combined hedge algebra fuzzy logic for target tracking and obstacle avoidance, whereas Dang et al. \cite{van2022collision} integrated a Soar-based cognitive agent with the Dynamic Window Approach (DWA) for safe human following. Ye et al. \cite{11106254} further proposed a human-following framework with target re-identification, enabling the robot to re-acquire the user after temporary target loss. More recently, Wang et al. \cite{11184399} introduced a vision-based human-following system using human pose estimation. However, due to the limited field of view (FoV) of the vision sensor, these approaches are largely restricted to following the user from behind and remain vulnerable to target occlusions.

Although the following-from-behind approach has achieved considerable success and has been extensively adopted, it presents two key limitations. First, it can cause discomfort and uncertainty for the individual being followed, as they may frequently check the robot's position, leading to distraction. Second, the robot’s position behind the person complicates interaction and collaboration, particularly when there is a need to place or retrieve objects from the robot. Furthermore, researches on human walking behavior \cite{repiso2020adaptive, jafari2024pedestrians} has shown that when robots interact with humans, they remain within the individual’s field of view, which promotes higher levels of social acceptance and makes the interaction more comfortable for the human.

Recent studies have broadened the application of tracking controllers to create robots capable of accompanying humans by following side-by-side, thereby enhancing natural and comfortable human-robot interactions. Several studies have integrated advanced controllers like MPC \cite{peng2023mpc} and Linear Quadratic Regulator (LQR) \cite{su2024lqr} into human-robot interaction models, enabling precise control over position, orientation, and relative velocity, while simulating human-like obstacle avoidance. However, these approaches often neglect psychological comfort, especially in maintaining appropriate social distances. To address this, Peng et al. \cite{peng2024dual} proposed a dual-loop closed control system: the outer-loop MPC controller ensures precise posture control while tracking the target's speed and direction, while the inner-loop impedance controller adjusts interaction forces to maintain a comfortable distance. Additionally, Repiso et al. \cite{repiso2017line} introduced an Extended Social Force Model (SFM) to maintain angle and distance during shared movement. Nonetheless, real-world scenarios often present spatial limitations, making it impractical for the robot to consistently move side-by-side with the human. Consequently, control systems should be adaptable to specific applications and environmental constraints, enabling the robot to follow behind or move alongside as needed \cite{islam2019person}.
\begin{table*}[t]
\caption{Comparison with Related Works}
\centering
\begin{tblr}{
  width=\linewidth,
  colspec={X[2.85,l,m] X[1.5,c,m] X[1.5,c,m] X[0.85,c,m] X[1.3,c,m] X[0.85,c,m] X[1.1,c,m] X[0.85,c,m] X[0.8,c,m]},
  rowsep=0.23em,
  row{2}={bg=gray!20}
}
\hline[1.35pt]\hline
\textbf{Method} & \textbf{Predefined Tracking Point} & \textbf{Adaptive Companionship} & \textbf{Follow Behind} & \textbf{Follow Side-by-Side} & \textbf{Follow In-Front} & \textbf{Obstacle Avoidance}  & \textbf{Social Space} & \textbf{Outdoor}  \\
\hline[1.0pt]
\textbf{Proposed} & \textbf{No} & \textbf{Yes} & \textbf{Yes} & \textbf{Yes} & \textbf{Yes} & \textbf{Yes} & \textbf{Yes} & \textbf{Yes}  \\
Corridor-Intersection Recognition \cite{yao2021laser} & Yes & No & Yes  & Only in corridors & No & No &No & - \\
PID \cite{chen2019human}, Gait Recog. \cite{7852457}, Deep-MPC \cite{10606944}, HFC \cite{11184399} & Yes & No & Yes & No & No & No &No & -  \\
Hedge Algebra Fuzzy \cite{van2023human}, DWA \cite{van2022collision}, RPF-Search\cite{11106254}, Deep-RL \cite{pang2020efficient}  & Yes & No & Yes & No & No & Yes &No & -  \\
Adaptive SFM \cite{repiso2020adaptive}, MPC \cite{peng2023mpc}, LQR \cite{su2024lqr}, Extended SFM \cite{repiso2017line} & Yes & No & No & Yes & No & Yes &No & - \\
Dual Closed-Loop \cite{peng2024dual} & Yes & No & No & Yes & No & Yes &Yes & - \\
Finite-Time Control \cite{yan2021human}, NMPC \cite{sekiguchi2021uncertainty} & Yes & No & No & No & Yes & No &No & No  \\
IMU-Laser \cite{cifuentes2014human} & Yes & No & No & No & Yes & No &No & Yes  \\
LBGP \cite{nikdel2021lbgp}, MCTS-DRL \cite{leisiazar2023mcts}, LSTM-DRL \cite{10869380} & Yes & No & No & No & Yes & Yes &No & No  \\
\hline\hline[1.35pt]
\end{tblr}\label{table:compare}
\end{table*}

The ability to follow a human from the front is an important topic in human–robot interaction, with growing research interest. However, accurately modeling the stochastic nature of human motion in front of a robot remains challenging, particularly in complex environments such as T-junctions. To address this, Moustris et al. \cite{moustris2016intention} proposed an intention recognition method to improve indoor assistance through dynamic local planning. Sekiguchi et al. \cite{sekiguchi2021uncertainty} introduced a nonlinear model predictive controller for front-following that accounts for uncertainty in human motion, enabling smoother and more adaptive behavior. More recently, reinforcement learning (RL) has been applied to improve obstacle avoidance and navigation in unknown environments via trial-and-error learning \cite{11353266}, and extended to human–robot interaction tasks, including human tracking and adaptation to dynamic behaviors \cite{11180937,10966209,nikdel2021lbgp,leisiazar2023mcts,10869380}. Despite these advances, accurately estimating human motion remains difficult, even for learning-based methods, which can degrade navigation performance, hinder human movement, and raise safety concerns. Moreover, many existing approaches neglect environmental constraints and focus primarily on front-tracking in static or simplified settings, limiting their real-world applicability.

In real-world scenarios, spatial constraints often hinder robots from maintaining fixed tracking positions, such as consistently walking beside, behind, or in front of a person. To address these challenges, robots should be capable of dynamically adjusting their tracking positions in response to changes in the surrounding environment, rather than relying on a single predefined strategy. However, few studies have addressed this adaptive behavior. Yao et al. \cite{yao2021laser} introduced a corridor-intersection recognition module, enabling robots to transition from following behind to walking alongside during corner navigation. Although this method utilizes spatial features to change the following position, the algorithm only operates at corridor intersections, resulting in rigid transitions and unstable trajectories. Therefore, it is not well suited to real-world environments, where flexible and adaptive tracking strategies are essential for safe and effective robot behavior.

In human-companion robotics, significant progress has been made in both classical control and learning-based navigation methods. Classical approaches provide strong safety guarantees and accurate dynamic modeling but often lack adaptability in dynamic environments. In contrast, learning-based methods enable adaptive decision-making without requiring an explicit system model. However, they suffer from the sim-to-real gap and lack inherent safety guarantees for low-level control, a critical limitation in human–robot interaction \cite{10675394}. Although advanced architectures, such as Bi-GRU \cite{MONTERO2025101942} and Transformers \cite{10007923}, have significantly improved general navigation performance, more recently, vision-language model (VLM)-based navigation strategies have also been explored for group-following tasks \cite{vu2026adaptive}. Nevertheless, their deployment in human-companion robotics remains challenging because they must simultaneously achieve real-time performance, maintain tracking accuracy, guarantee user safety, and adapt to rapid changes in human motion.
\begin{figure}[t]
\centerline{\includegraphics[scale=0.63,page=1]{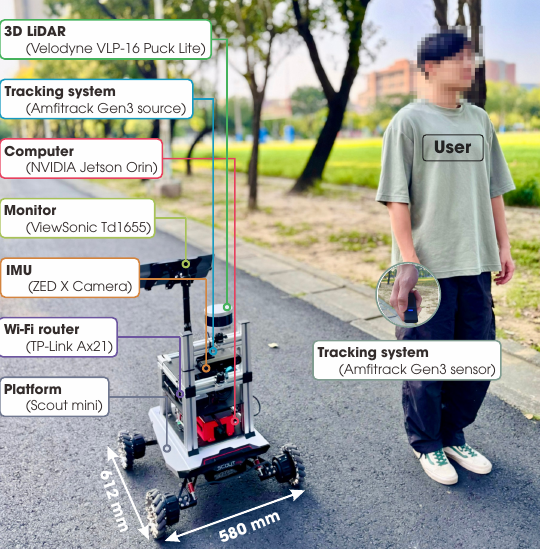}}
\caption{Overview of the human-interactive robot.}
\label{fig:system}
\end{figure}

In this paper, we propose an adaptive human-accompanying approach that accounts for workspace context and social acceptance, rather than relying on fixed tracking positions as in prior human-following methods. As summarized in Table \ref{table:compare}, the proposed method enables the robot to dynamically adjust its tracking position according to the environment. Specifically, we define an interactive zone around the human within the workspace and integrate it with a designed objective function to construct the input for an RL model. The trained policy outputs an optimized tracking position. A Model Predictive Path Integral (MPPI) controller combined with Control Barrier Functions (CBF) is then employed for execution, ensuring safety, improving social compliance, and enabling obstacle avoidance during human accompaniment. The main contributions of this research are as follows:
\begin{enumerate}
    \item An adaptive human-accompanying strategy is proposed to dynamically regulate the robot's tracking position according to the surrounding environment, rather than relying on fixed relative positions.
    \item An interactive space based on an elliptical human-occupancy model is introduced as the state representation for a reinforcement learning (RL) policy, together with a newly designed reward function.
    \item An MPPI-CBF controller is developed by incorporating proxemics and interactive-space constraints to ensure safe and socially compliant navigation.
    \item The proposed framework is validated on an omnidirectional mecanum-wheeled robot in both indoor and outdoor environments without prior map information, demonstrating robust performance in real-world environments beyond the static settings considered in prior work.
    \item Compared with representative state-of-the-art methods \cite{chen2019human, van2022collision, peng2023mpc, pang2020efficient}, the proposed framework achieves up to 24\% higher success rate and 47\% better tracking accuracy, demonstrating its effectiveness and robustness in real-world environments.
\end{enumerate}

The structure of this paper is as follows: Section \ref{sec:system} outlines the system design and problem formulation of the human-accompanying robot. Section \ref{sec:goal} explains the human-robot interactive space, which is then used in the RL network in Section \ref{sec:rl}. Section \ref{sec:mppi_cbf} introduces the human-companion control algorithm, incorporating the MPPI controller and CBF. Experimental setup, results, and discussions are addressed in Section \ref{sec:results}. Finally, Section \ref{sec:conclusion} summarizes key findings and future work.
\section{System Design and Problem Formulation}\label{sec:system}
In this section, we present the design and hardware components of the human-companion robot and the problem formulation for the human-accompanying task.
\subsection{Design of the Human-Companion Robot}
Fig.~\ref{fig:system} illustrates the configuration of the human–companion robot system. The platform is equipped with a Velodyne VLP-16 Puck Lite 3D LiDAR for environmental perception and an Amfitrack Gen 3 electromagnetic tracker \cite{amfitrack_guide} for human tracking. A source mounted on the robot and a sensor attached to the user enable reliable target identification while preventing confusion with nearby pedestrians. The tracker provides six-degrees-of-freedom relative pose measurements and remains effective under visual occlusion, overcoming limitations of camera- and LiDAR-based approaches \cite{van2023human, chen2019human, yao2021laser}. Its onboard IMU, fused through an Extended Kalman Filter (EKF), further improves tracking accuracy and robustness. Measurement confidence is monitored in real time, allowing the robot to stop safely if the signal quality falls below a predefined threshold. Compared with UWB- and RF-based systems \cite{11127912, van2022collision}, which estimate only user position, Amfitrack Gen 3 reliably estimates both position and orientation.

 The proposed system supports flexible operation across multiple environments without being limited to a fixed setup, in contrast to prior work relying on Motion Capture (MoCap) systems for human tracking \cite{nikdel2021lbgp, leisiazar2023mcts, sekiguchi2021uncertainty}. It integrates an IMU-equipped ZED X camera from Stereolabs to estimate robot velocity and acceleration, while all sensor data are processed on an NVIDIA Jetson Orin using CUDA acceleration under Ubuntu 22.04 with JetPack 6.1 and ROS2 Humble. The platform is built on an Agilex Scout Mini with four mecanum wheels, enabling omnidirectional motion for dynamic human accompaniment.

 Based on the proposed system configuration, the IMU operates at 200 Hz, while the user position obtained from the Amfitrack system is provided at 120 Hz. The robot odometry is updated at 50 Hz, and the Velodyne VLP-16 Puck Lite LiDAR provides measurements at 10 Hz. For the proposed companionship strategy, the RL-based policy operates at 30 Hz, as described in Section~\ref{sec:rl}, while the tracking controller, also presented in Section~\ref{sec:mppi_cbf}, operates at 100 Hz.
\subsection{Problem Formulation}
\begin{figure}[t]
\centerline{\includegraphics[scale=0.37,page=1]{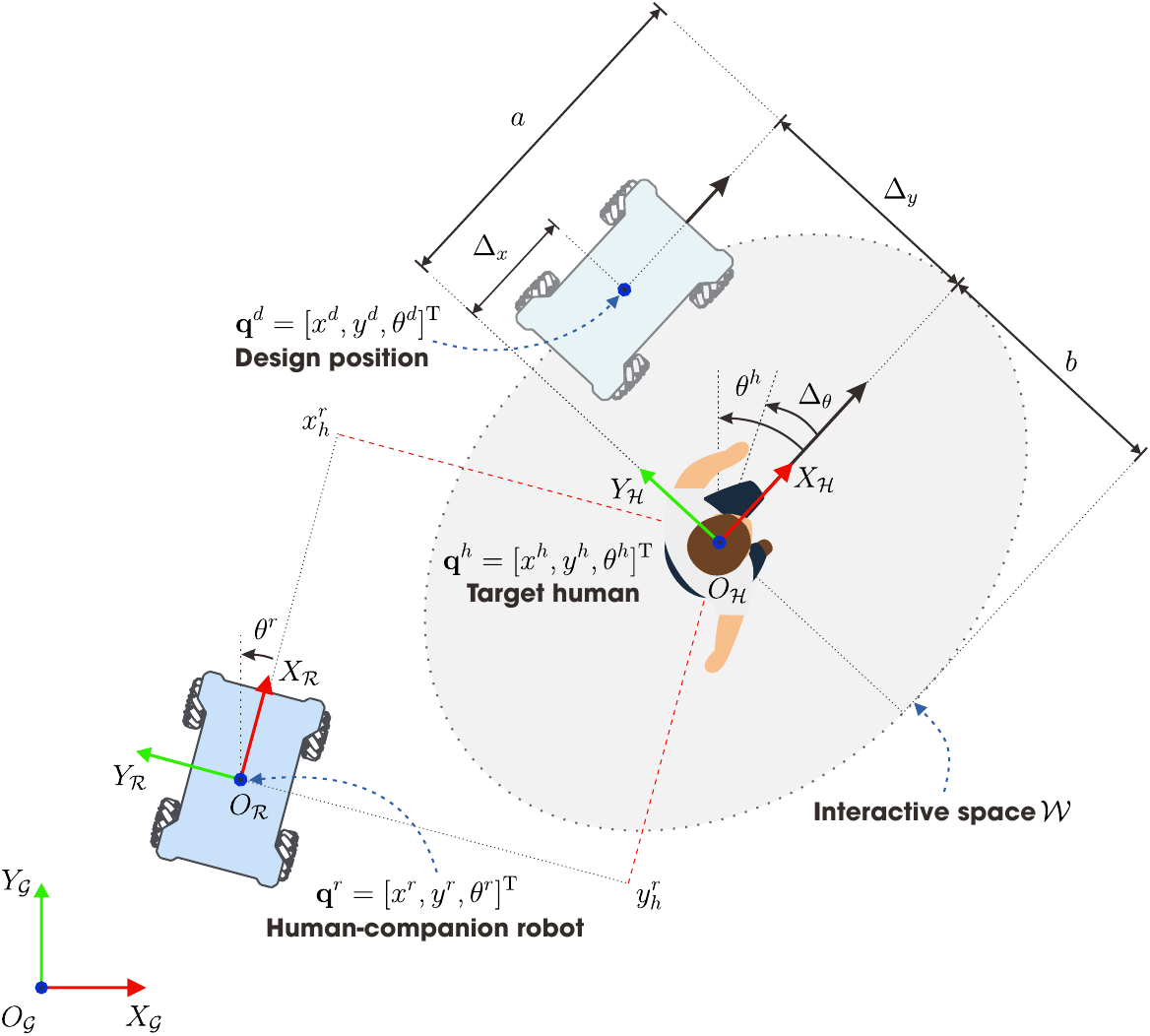}}
\caption{Schematic model and generalized coordinates.}
\label{fig:model}
\end{figure}

The schematic models and generalized coordinates of the proposed robot are illustrated in Fig.~\ref{fig:model}, with the kinematic model of a mecanum-wheeled omnidirectional robot derived and presented by \cite{muir1987kinematic}. The robot and human states, defined as $\mathbf{q}^r = [x^r, y^r, \theta^r]^\text{T} \in \mathbb{R}^3$ and $\mathbf{q}^h = [x^h, y^h, \theta^h]^\text{T} \in \mathbb{R}^3$, represent their positions and orientations in the global frame $\mathcal{G}$, respectively. The human’s position and orientation relative to the robot’s local frame $\mathcal{R}$ is given by $\mathbf{q}^r_h = [x^r_h, y^r_h, \theta^r_h]^\text{T} \in \mathbb{R}^3$. The transformation of coordinates between the global and local frames for the human’s state can be expressed as:
\begin{align}
    \mathbf{q}^h = {^{\mathrm{G}}_{\mathrm{R}}{\mathbf{T}}}\mathbf{q}^r_h
\end{align}
where
\begin{align*}
   {^{\mathrm{G}}_{\mathrm{R}}{\mathbf{T}}} = \begin{bmatrix}
        \mathbf{J}(\theta^r) & \mathbf{q}^r \\
        0 & 1
    \end{bmatrix}, \quad
    \mathbf{J}(\theta^r) = \begin{bmatrix}
        \cos\theta^r & -\sin\theta^r & 0 \\
        \sin\theta^r & \cos\theta^r & 0 \\
        0 & 0 & 1
    \end{bmatrix},
\end{align*}
$^{\mathrm{G}}_{\mathrm{R}} \mathbf{T}$ is the transformation matrix from $\mathcal{R}$ to $\mathcal{G}$, and $\mathbf{J}(\theta_r)$ is the rotation transformation matrix.

In the human-accompanying task, the robot should maintain the desired position $\mathbf{q}^d = [x^d, y^d, \theta^d]^\text{T} \in \mathbb{R}^3$ in the global frame, specifically ensuring that $\lim_{t \rightarrow \infty} \left| \mathbf{q}^r - \mathbf{q}^d \right| = 0$ by adjusting the relative position $(x^r_h, y^r_h)$ between the robot and the human target, while also ensuring that the robot can follow the orientation $\theta^r_h$. Thus, the control objective can be expressed as follows:
\begin{align}
    x^r_h = -\Delta_x, \quad y^r_h = -\Delta_y, \quad \theta^r_h = -\Delta_{\theta},
\end{align}
where $\Delta_x$ and $\Delta_y$ represent the relative distances along the $x$ and $y$ axes from the desired position to the human, respectively, and $\Delta_\theta$ denotes the orientation between the human and the robot in the local coordinate frame $\mathcal{R}$.
\begin{remark}
Considering fixed the desired tracking position $\mathbf{q}^d$ in prior studies is often impractical in real-world scenarios due to the limited workspace and the presence of obstacles within the operational environment. This study proposes an adaptive companion algorithm that enables the robot to autonomously adjust its position relative to the human, such as in front, beside, or behind while ensuring it remains within the designated interactive space and actively avoids collisions.
\end{remark}
\section{Human-Robot Interactive Space}\label{sec:goal}
This section introduces a collision-free interactive space and cost functions for selecting the desired position. 
\subsection{Collision-Free Interactive Space}
The interactive space $\mathcal{W}$ is defined as the set of $n$ design points $\mathbf{q}^r_{d,i}=[x^r_{d,i}, y^r_{d,i}, \theta^r_{d,i}]^\text{T} \in \mathbb{R}^3$ for $i = 1, \ldots, n$ in the frame $\mathcal{R}$. These points are located within an ellipse with a semi-major axis $a$ and a semi-minor axis $b$, centered at the human position. An elliptical representation is adopted to model human occupancy, reflecting the anisotropic characteristics of pedestrian locomotion, wherein humans generally require a larger frontal space than lateral space to maintain forward motion, as noted in the Social Force Model \cite{9716053,helbing1995social}. Thus, the interactive space is given as:
\begin{align}
\mathcal{W} = \left\{ \mathbf{q}^r_{d,i} \mid i = 1, \ldots, n \right\}, \quad \mathbf{q}^r_{d,i} = {^{\mathrm{R}}_{\mathrm{H}}{\mathbf{T}}}\mathbf{q}^h_{d,i},
\end{align}
where
\begin{align*}
   {^{\mathrm{R}}_{\mathrm{H}}{\mathbf{T}}} = \begin{bmatrix}
        \mathbf{J}(\theta^r_h) & \mathbf{q}^r_h \\
        0 & 1
    \end{bmatrix}, \quad
    \mathbf{J}(\theta^r_h) = \begin{bmatrix}
        \cos\theta^r_h & -\sin\theta^r_h & 0 \\
        \sin\theta^r_h & \cos\theta^r_h & 0 \\
        0 & 0 & 1
    \end{bmatrix},
\end{align*}
$^{\mathrm{R}}_{\mathrm{H}} \mathbf{T}$ is the transformation matrix from $\mathcal{H}$ to $\mathcal{R}$, and $\mathbf{J}(\theta^r_h)$ is the rotation transformation matrix. The designed position in the interactive space, represented in the human coordinates $\mathcal{H}$, is given as:
\begin{align}
\mathbf{q}^h_{d,i} = \begin{bmatrix}
a \cos \phi_i \\
b \sin \phi_i \\
0
\end{bmatrix}, \quad  \phi_i = \frac{2\pi}{n} (i - 1),
\end{align}
where $\phi_i \in [0, 2\pi)$ is the angle corresponding to each position $i$-th in the interactive space $\mathcal{W}$.

The space $\mathcal{W}$ is refined by eliminating points that conflict with obstacles. To enhance computational efficiency, we utilize the spatio-temporal voxel layer costmap \cite{stvl}, which is generated from a 3D LiDAR sensor mounted on the robot's frame, instead of directly evaluating each point in $\mathcal{W}$ using raw point cloud data. We define $\mathcal{L}$ as the set of costs within the local costmap, where $l(\mathbf{q})$ denotes the function used to determine the cost at a given point $\mathbf{q}$. The collision-free interactive space is represented as $\mathcal{P} = \{ \mathcal{P}_i \}_{i=1}^k$, comprising $k$ points (with $k \leq n$), which is defined as follows:
\begin{align}
\mathcal{P} = \left\{ \mathbf{q}^r_{d,i} \in \mathcal{W} \mid 1 \leq i \leq n, \, l(\mathbf{q}^r_{d,i}) < R \right\},
\end{align}
where $R$ represents the safe threshold cost between the position of a given point and the nearest obstacle.
\begin{figure*}[t]
\centering
\includegraphics[scale=0.235,page=1]{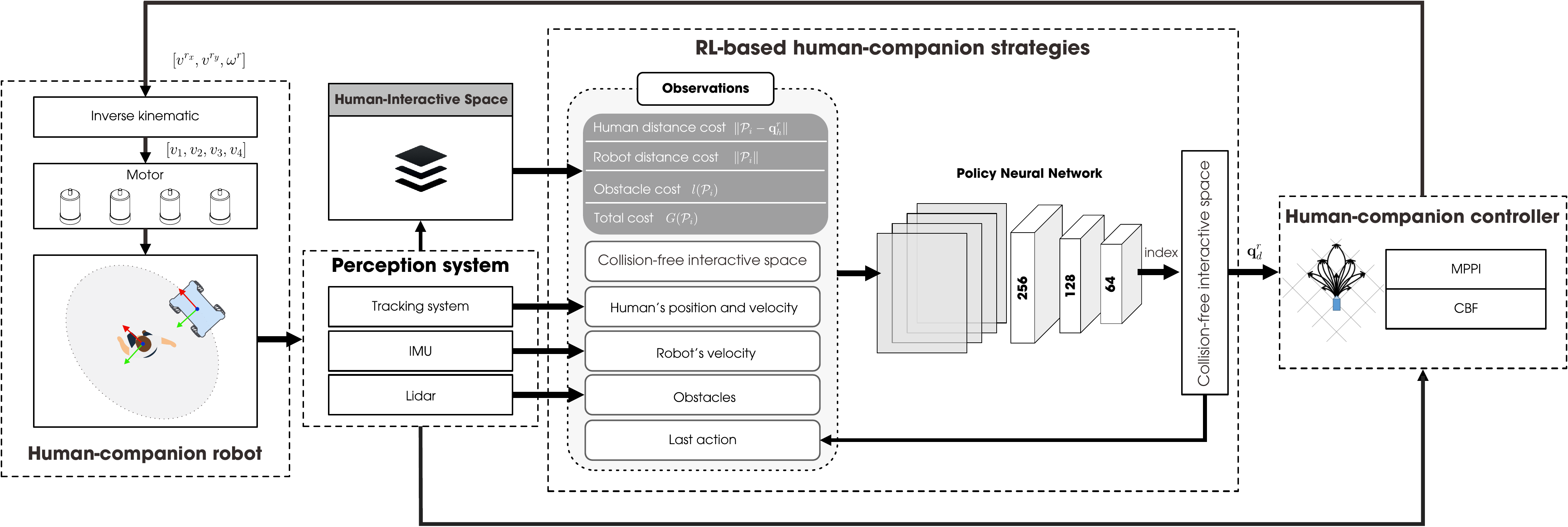}
\caption{The overall structure of the system architecture. The robot observes the proposed cost functions, total cost, collision-free space, robot velocity, human position and velocity, obstacles, and previous actions. Policy determines an action index within the collision-free space and then provides a target position for the human-companion controller.}
\label{fig:network}
\end{figure*}
\subsection{Objective Function}
After establishing the collision-free interactive space, we introduce an objective function designed to evaluate the priority of each following point based on three criteria as follows:
\begin{align}
G(\mathcal{P}_i) = \alpha \, \left\|\mathcal{P}_i-\mathbf{q}^r_h \right\| + \beta \, \left\|\mathcal{P}_i\right\| + \gamma \, l(\mathcal{P}_i) .
\end{align}

During navigation, the robot selects a tracking point close to the human to enhance interaction. To evaluate this, we define the cost function $\left|\mathcal{P}_i - \mathbf{q}^r_h\right|$. However, proximity to the human alone is insufficient, as it may lead to points far from the robot and result in unstable motion. Therefore, we introduce $\left|\mathcal{P}_i\right|$ to measure the distance from each candidate point to the robot, balancing human proximity and motion smoothness. To ensure obstacle avoidance, we also include $l(\mathcal{P}_i)$ from the local costmap. The weighting coefficients $\alpha$, $\beta$, and $\gamma$ balance these criteria, and the desired point is then selected as follows:
\begin{align}
\mathbf{q}^r_{d} = \operatorname*{argmin}_{i \in \{1, \ldots, k\}} G(\mathcal{P}_i).
\label{eq:objective}
\end{align}
\begin{remark}
By minimizing the objective function in \eqref{eq:objective} over each point in $\mathcal{P}$, a desired location satisfying the specified conditions under given weights can be obtained. However, these weights are typically environment-dependent, making manual tuning impractical in diverse real-world scenarios. To address this limitation, Section \ref{sec:rl} introduces an RL-based human-companion method that learns adaptive policies via reward evaluation, enabling the robot to autonomously select suitable companion positions in complex environments.
\end{remark}
\section{RL-Based Human-Accompanying Strategies}\label{sec:rl}
Unlike previous studies that used RL to follow the human by considering only the relative position between the human and the robot, we propose a novel approach that leverages the proposed interactive space as the state space for RL. Additionally, we design a new reward function specifically for the human-accompanying task and present the policy training process. The overall system architecture is illustrated in Fig.~\ref{fig:network}.
\subsection{State Space and Action Space}
The state space $\mathbf{s}_t$ for the human-companion task is informed by the relevant observational data required by the robot and the proposed following space. We define the state space at time $t$ as follows:
\begin{align}
\mathbf{s}_t = \left(\mathbf{s}^c_t, \mathbf{s}^p_t, \mathbf{s}^o_t, \mathbf{s}^r_t, \mathbf{s}^h_t, \mathbf{s}^a_t \right),
\end{align}
where $\mathbf{s}^c_t = \{\mathbf{s}^c_{t,i} \}_{i=1}^k$ captures essential cost functions along with the proposed objective function. Each component $\mathbf{s}^c_{t,i}$ is defined as follows:
\begin{align*} 
\mathbf{s}^c_{t,i} = \Big[ \left\|\mathcal{P}_i-\mathbf{q}^r_h \right\|, \left\|\mathcal{P}_i\right\|, l(\mathcal{P}_i) , G(\mathcal{P}_i)\Big] .
\end{align*}

The term $\mathbf{s}^p_t$ represents the collision-free interactive space $\mathcal{P}$, and $\mathbf{s}^o_t=\mathcal{L}$ denotes the observation space of obstacles surrounding the robot at each time step, which is represented by a local cost map. The robot’s current velocity is given by $\mathbf{s}^r_t = [\dot{\mathbf{q}_r}]$, while $\mathbf{s}^h_t = [\mathbf{q}^r_h, \dot{\mathbf{q}^r_h}]$ specifies the relative position and velocity of the human in relation to the robot. Additionally, $\mathbf{s}^a_t$ denotes the previous action.

The action is the index $i$  corresponding to a collision-free interactive point in the space $\mathcal{P}$, where the $i$-th tracking position in the action space is used as the desired point for the robot.
\subsection{Reward Functions}
Previous studies \cite{nikdel2021lbgp,kastner2022human,leisiazar2023mcts,pang2020efficient} on human-following tasks using RL have primarily focused on reward functions to maintain a specific distance between the robot and the user, often overlooking critical spatial factors such as obstacle avoidance, the robot’s repositioning capabilities, and the safe interaction zone between the user and the robot. In this study, we propose a reward function to address these factors:
\begin{align}
\mathbf{r}_t(\mathbf{s}_t) = \mathbf{r}^h_t + \mathbf{r}^a_t  + \mathbf{r}^c_t.
\end{align}

Distance to Human Reward ($\mathbf{r}^h_t$): To encourage the robot to move closer to the human, the robot receives a positive reward, denoted as $r_{\text{tracking}}$, whenever the distance between the robot and the human remains within a permissible limit~$D$. Conversely, a dynamic reward system based on the current~$\Vert \mathbf{q}_{h,t}^r \Vert$ and previous positions $\Vert \mathbf{q}_{h,t-1}^r \Vert$ is employed. The robot will receive a positive reward if the current distance to the human is less than the previous distance, and a negative reward if it moves further away. The reward structure is defined as follows:
\begin{align}
\mathbf{r}_t^h = 
\begin{cases} 
\sigma_h \left( \Vert \mathbf{q}_{h,t-1}^r \Vert -\Vert \mathbf{q}_{h,t}^r \Vert \right) & \text{if } \Vert \mathbf{q}_{h,t}^r \Vert > D \\ 
r_{\text{tracking}} & \text{otherwise}
\end{cases},
\end{align}
where, $\sigma_h$ represents the weight assigned to the reward function.

Action Smoothness Reward ($\mathbf{r}^a_t$): The action reward function is employed to enhance the smoothness of the trajectory by limiting abrupt positional transitions. This approach stabilizes the robot's path by encouraging continuous adjustments to its actions. The action reward function is defined as follows:
\begin{align}
\mathbf{r}_t^a = -\sigma_a \left | \mathbf{a}_t - \mathbf{a}_{t-1} \right|,
\end{align}
where $\sigma_a$ is the weight applied to the action reward, and $\mathbf{a}_t$ and $\mathbf{a}_{t-1}$ represent the robot's actions represent the robot's actions at time $t$ and time $t-1$, respectively.

Collision Avoidance Penalty Reward ($\mathbf{r}^c_t$): Although the space $\mathcal{P}$ has been cleared of collision points, operating in dynamic or crowded environments necessitates proactive collision avoidance measures before conflicts arise. This penalty is defined as follows:
\begin{align}
\mathbf{r}^c_t =
\begin{cases}
r_{\text{collision}} & \text{if } \min(\mathcal{L}) \ge  R \\ 
0 & \text{otherwise}
\end{cases},
\end{align}
where $r_{\text{collision}}$ denotes the penalty incurred by the robot in the event that the minimum distance to an obstacle, as determined by Lidar, is less than the safety threshold $R$.

\begin{figure}[t]
\centering
 \includegraphics[scale=0.13,page=1]{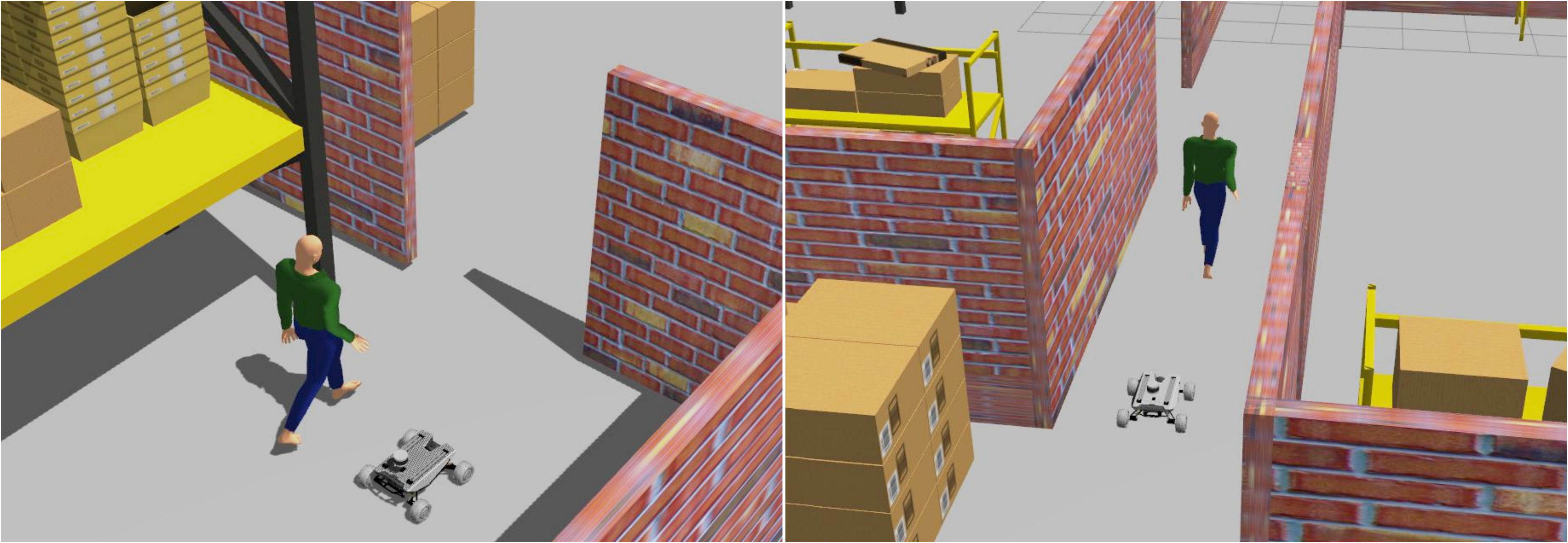}
\caption{Illustration of the training environment.}
\label{fig:env_traning}
\end{figure}
The total reward is formulated with components that exhibit strong coupling relationships. Specifically, a direct competition exists between the distance-to-human reward $r_t^h$ and the action smoothness reward $r_t^a$. While $r_t^h$ continuously drives the robot to minimize tracking error through immediate tactical adjustments, $r_t^a$ penalizes large or abrupt changes in the action space to ensure trajectory stability. Overemphasizing $r_t^h$ yields a highly responsive but oscillatory and jerky motion profile, whereas overweighting $r_t^a$ results in smooth but sluggish behavior that fails to adapt to sudden changes in human direction. Crucially, the collision penalty $r_t^c$ acts as an absolute safety override that dominates this trade-off framework. By assigning a significantly larger magnitude to the collision penalty ($r_{collision} = -200$) relative to the maximum tracking reward ($r_{tracking} = 50$), the policy strictly prioritizes safety over all other criteria. When in close proximity to obstacles, the policy proactively sacrifices tracking accuracy or executes abrupt maneuvers that reduce smoothness, enforcing a clear hierarchical priority in which safe navigation strictly supersedes both operational performance and user comfort.
\subsection{Policy Training}
The policy was trained on a workstation equipped with an Intel Xeon W5-3435X CPU and an NVIDIA A6000 GPU using the PPO algorithm implemented in the skrl library \cite{serrano2023skrl} in a simulation environment as shown in Fig.~\ref{fig:env_traning}. The actor network comprises three fully connected layers with 256, 128, and 64 neurons, followed by an output layer matching the discrete action space dimension. Reward parameters were set empirically as $n=64$, $a=1$, $b=0.7$, $D=0.75$, and $R=30$ to balance tracking performance and safety, ensuring stable convergence. Observations were normalized to ensure consistent input scaling between simulated and real sensor data. Domain randomization was applied to human walking speed, lateral deviation, sensor noise, agent-relative positioning, and obstacle placement, enabling the policy to be trained on a wider range of scenarios. In addition, to enable the policy to adapt to different interaction-space sizes, the dimensions of the interaction space were also randomized.

\begin{figure}[t]
\centering
 \includegraphics[scale=0.24,page=1]{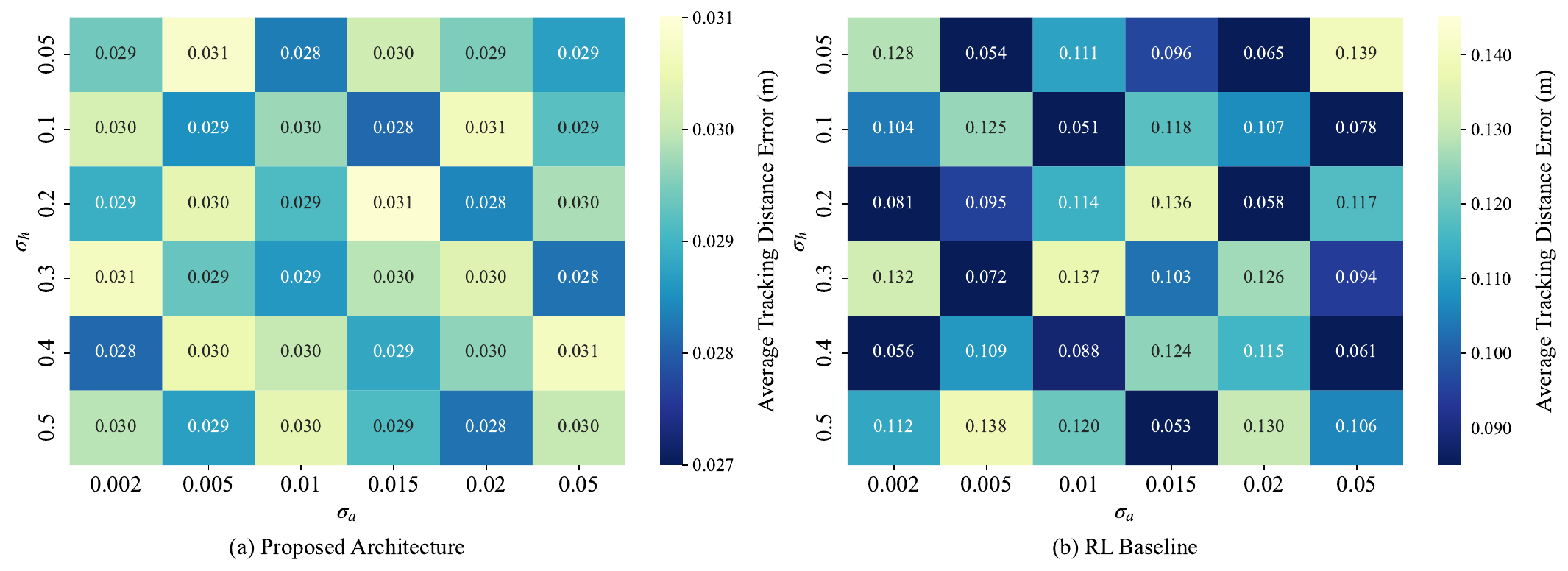}
\caption{Policy performance under varying hyperparameter configurations: (a) Proposed architecture and (b) RL baseline.}
\label{fig:parameter_sensitivity_comparison_matrix}
\end{figure}
The system employs a hierarchical control architecture in which the RL policy functions as a high-level planner that selects the optimal tracking position within the interactive space, while low-level execution is handled by the MPPI controller (Section~\ref{sec:mppi_cbf}). This separation significantly reduces the policy’s sensitivity to specific network architectures and reward parameters, since the RL policy focuses solely on selecting the target position, whereas the MPPI controller compensates for variations in physical dynamics and environmental perturbations to drive the robot toward the selected position. The policy was validated during training by varying the parameters $\sigma_h$ and $\sigma_a$. The results in Fig.~\ref{fig:parameter_sensitivity_comparison_matrix} show that the proposed architecture is less sensitive to these parameters, whereas the baseline RL method exhibits significant performance degradation. Moreover, by generating high-level waypoints in the interactive space instead of issuing direct motor commands, the system decouples planning from uncertainties in physical dynamics, such as wheel slippage or uneven terrain.

In this architecture, the high-level DRL policy operates at a lower frequency of 30 Hz, focusing on strategic decision-making for selecting appropriate accompanying positions within the interaction space. By contrast, the low-level MPPI controller runs at a higher frequency of 100 Hz to ensure real-time trajectory tracking and fast dynamic compensation. The higher operational rate of the MPPI layer is essential for handling rapid system dynamics, mitigating disturbances, and ensuring smooth and stable execution of high-level commands. This also helps prevent oscillatory behavior that may arise when the reference position is updated faster than the system can reliably track.
\section{Human-Companion Controller}\label{sec:mppi_cbf}
In this section, we introduce a human-companion controller based on MPPI \cite{williams2018information}, which integrates a CBF, adapted from \cite{yin2023shield}, as a safety mechanism to prevent the robot from entering the human’s intimate zone, thereby enhancing user safety and comfort.
\subsection{MPPI-Based Human Accompanying}
The discrete state of the human-companion robot is given by:
\begin{align}
   \mathbf{q}_{k+1}^r = \mathbf{q}_{k}^r + \mathbf{J}\big(\theta^r_k\big) \, \mathbf{v}^r_k \, \Delta t,
\end{align}
where $\mathbf{q}^r_k$ denotes the robot's state, and $\mathbf{v}^r_k = [v^{r_x}_k, v^{r_y}_k, \omega^r_k]^\text{T} \in \mathbb{R}^3$ is the control input applied at time step $k$, for $k = 0, \dots, K-1$ with a sampling period $\Delta t$. Here, $v^{r_x}_k$, $v^{r_y}_k$, and $\omega^r_k$ represent the robot's linear velocities along the $x$-axis and $y$-axis, and its angular velocity in $\mathcal{R}$, respectively.

The control input $ \mathbf{v}^r_k $ is modeled as a Gaussian distribution, given by $ \mathbf{v}^r_k \sim \mathcal{N}(\mathbf{u}_k, \Sigma_\epsilon) $, where $ \mathbf{u}_k \in \mathbb{R}^3 $ is the mean and $ \Sigma_\epsilon \in \mathbb{R}^{3 \times 3} $ represents the covariance matrix. Consequently, the control sequence $ \mathbf{v} = (\mathbf{v}^r_0, \dots, \mathbf{v}^r_{K-1}) $ adheres to the distribution $ \mathbb{Q} $, with an associated probability density function given by:
\begin{align}
    q(\mathbf{v}) = Z \prod_{k=0}^{K-1} \exp\left( -\frac{1}{2} (\mathbf{v}^r_k - \mathbf{u}_k)^\text{T} \Sigma_\epsilon^{-1} (\mathbf{v}^r_k - \mathbf{u}_k) \right),\label{eq:(qV)}
\end{align}
where $Z = \left( (2\pi)^3 |\Sigma_\epsilon| \right)^{-\frac{1}{2}}$. The objective function for the human-accompanying task is defined as:
\begin{align}
    J(\mathbf{u})= \mathbb{E}_\mathbb{Q} \left[ \phi( \mathbf{q}^r_{K}) + \sum_{k=0}^{K-1} \left( c(\mathbf{q}^r_{k}) + \frac{\lambda}{2} \mathbf{u}_k^\text{T} \Sigma_\epsilon^{-1} \mathbf{u}_k \right) \right],\label{eq:optimal}
\end{align}
where $ c(\mathbf{q}^r_{k}) $ is the stage cost, and $ \phi( \mathbf{q}^r_{K}) $ is the terminal cost. The optimal distribution $ \mathbb{Q}^* $ that minimizes this objective has a density function given by:
\begin{align}
    q^*(\mathbf{v})= \frac{1}{\mu} \exp\left( -\frac{1}{\lambda} \left( \phi( \mathbf{q}^r_{K}) + \sum_{k=0}^{K-1} c(\mathbf{q}^r_{k}) \right) \right) p(\mathbf{v}),\label{eq:(q*V)}
\end{align}
where $ p(\mathbf{v}) $ is the density function of an uncontrolled base distribution $ \mathbb{P} $, which corresponds to a zero-mean control sequence ($ \mathbf{u} = 0 $) and
\begin{align}
    \mu = \int \exp\left(-\frac{1}{\lambda} \left( \phi( \mathbf{q}^r_{K}) + \sum_{k=0}^{K-1} c(\mathbf{q}^r_{k}) \right)\right) \, p(\mathbf{v}) \, \mathrm{d}\mathbf{v}.
\end{align}

The optimization problem \eqref{eq:optimal} can be reformulated as minimizing the Kullback-Leibler divergence between the associated probability distributions in \eqref{eq:(qV)} and \eqref{eq:(q*V)}. By applying the technique of importance sampling, the optimal control mean $\mathbf{u}_k^+$ is obtained as:
\begin{align}
    \mathbf{u}_k^+ = \mathbb{E}_\mathbb{Q}\left[\mathbf{v}^r_k \, w(\mathbf{v})\right], \quad w(\mathbf{v}) = \frac{1}{\eta} \exp\left(-\frac{1}{\lambda} S(\mathbf{v})\right),
    \label{eq:u}
\end{align}
where $\eta$ is given by:
\begin{align}
    \eta = \int \exp\left(-\frac{1}{\lambda} S(\mathbf{v})\right) \, \text{d}\mathbf{v}.
\end{align}

The trajectory cost $S(\mathbf{v})$ is defined as:
\begin{align}
    S(\mathbf{v}) = \phi( \mathbf{q}^r_{K}) + \sum_{k=0}^{K-1} c(\mathbf{q}^r_{k}) + \lambda \sum_{k=0}^{K-1} \mathbf{u}_k^\text{T} \Sigma_\epsilon^{-1} \mathbf{v}^k_{r}.
\end{align}

By employing Monte Carlo sampling for the control sequence in \eqref{eq:u}, with $\epsilon_k^m \sim \mathcal{N}(0, \Sigma_\epsilon)$ represents the sampled control noise for the $m$-th trajectory, for $m = 1, \dots, M$ at the $k$-th time step, the control input is updated as $\mathbf{v}^r_k = \mathbf{u}_k + \epsilon_k^m$. The control update law is then given by:
\begin{align}
     \mathbf{u}_k^+ = \mathbb{E}_\mathbb{Q}\left[( \mathbf{u}_k + \epsilon_k)w(\mathbf{v})\right]\approx \mathbf{u}_k + \frac{\sum^{M}_{m=1} \omega_{k}^{m} \epsilon_{k}^{m}}{\sum^{M}_{m=1} \omega_{k}^{m}},
\end{align}
where \( \omega_k^m \) denotes the weight associated with the control noise \( \epsilon_k^m \), defined as:
\begin{align}
    \omega^m = \exp \left(-\frac{1}{\lambda} (S^m - \min_{m=1,\dots,M} S^m ) \right),
\end{align}
where $\lambda$ is the hyperparameter that controls the selectivity of the MPPI algorithm in terms of the sampled trajectories. Let $S^m$ represent the cost associated with the $m$-th simulated trajectory, and $\min_{m=1, \dots, M} S^m$ ensures numerical stability without altering the optimal solution. Consequently, the cost of the $m$-th trajectory sample is given by:
\begin{align}
    S^m= \phi(\mathbf{q}^{r,m}_{K})+ \sum_{k=0}^{K-1} c(\mathbf{q}^{r,m}_{k}) + \lambda (\mathbf{u}_k^m)^{\text{T}}  \Sigma_\epsilon^{-1} (\mathbf{u}_k^m+ \epsilon^m_k).
\end{align}

The cost function for the human-accompanying task is formulated based on the desired tracking position $\mathbf{q}^{d}$ and the requirement for collision-free movement, as follows:
\begin{align}
c(\mathbf{q}^{r,m}_{k}) &= (\mathbf{q}^{r,m}_{k} - \mathbf{q}^{d})^\text{T} Q (\mathbf{q}^{r,m}_{k} - \mathbf{q}^{d}) + c_{\text{obs}},\label{eq:stage_cost}
\end{align}
where $Q = \operatorname{diag}(c_x, c_y, c_{\theta})$ is the cost weighting matrix that scales the cost for each state dimension, including position and orientation. The term $c_{\text{obs}}$ represents the obstacle avoidance cost and is defined as:
\begin{align}
c_{\text{obs}} = 
\begin{cases} 
c_{\text{collision}} & l(\mathbf{q}^{r,m}_{k}) \ge  R  \\
0 & \text{otherwise}.
\end{cases}.
\end{align}
\subsection{CBF-Based Safe Human-Robot Interaction}
While prior studies primarily focused on point-to-point tracking controllers without considering the safety and comfort essential for human-companion scenarios, we introduce a discrete-time control barrier function (DCBF) as an additional cost term to ensure the robot maintains a designated interactive space around the human while satisfying safety constraints. The robot’s state $\mathbf{q}^r$ is considered safe if it does not intersect with the region surrounding the human. Accordingly, the safe set $\mathcal{S} \subseteq \mathcal{D} \subset \mathbb{R}^3$ with respect to the human is defined as:
\begin{align}
\mathcal{S} = \left\{ \mathbf{q}^r \in \mathcal{D} \mid h(\mathbf{q}^{r}) \geq 0 \right\}.
\end{align}

It has been demonstrated in \cite{zeng2021safety} that the function $h(\mathbf{q}^{r})$ qualifies as the DCBF if it satisfies the condition:
\begin{align}
\Delta h(\mathbf{q}^{r}_k, \mathbf{v}^{r}_k) &\geq  -\delta h(\mathbf{q}^{r}_k),
\label{eq:condition}
\end{align}
where $\Delta h(\mathbf{q}^{r}_{k}, \mathbf{v}^{r}_{k}) = h(\mathbf{q}^{r}_{k+1}) - h(\mathbf{q}^{r}_{k})$ and $0 < \delta \leq 1$ ensures that the safety condition is maintained across time steps.

We select a barrier function to define the safety constraint and the safe region, shaped as an elliptical interactive space around the human, as follows:
\begin{align}
h(\mathbf{q}^{r}) = \frac{{(x_r^h)}^2}{a^2} + \frac{{(y_r^h)}^2}{b^2} - 1,
\end{align}
where $x_r^h$ and $y_r^h$ are determined through the coordinate transformation from the robot's coordinate system to the user’s coordinate system.
\begin{align}
    {^{\mathrm{H}}_{\mathrm{R}}{\mathbf{T}}} = {^{\mathrm{G}}_{\mathrm{H}}{\mathbf{T}}}^{-1}{^{\mathrm{G}}_{\mathrm{R}}{\mathbf{T}}}.
\end{align}
The cost function for MPPI with the DCBF is defined as:
\begin{align}
c_\text{cbf}(\mathbf{q}^{r}_k) = c_\text{safe} \max \left\lbrace (1-\delta) h\left(\mathbf{q}^{r}_{k}\right)-h\left(\mathbf{q}^{r}_{k+1}\right), 0\right\rbrace.
\end{align}
It is assumed that at the initial position, $\mathbf{q}^{r}_{0} = \mathbf{q}^{r}_{-1}$. If the condition given in \eqref{eq:condition} for the barrier function is satisfied, the cost function for the DCBF will be zero, ensuring that the robot maintains a safe distance from the human. Otherwise, the robot incurs a penalty weighted by $c_\text{safe}$. Therefore, the cost function defined in \eqref{eq:stage_cost} is redefined as:
\begin{equation}
\begin{split}
c(\mathbf{q}^{r,m}_{k}) = &\ (\mathbf{q}^{r,m}_{k} - \mathbf{q}^{d})^\text{T} Q (\mathbf{q}^{r,m}_{k} - \mathbf{q}^{d})+ c_{\text{obs}} + c_\text{cbf}(\mathbf{q}^{r,m}_k).
\end{split}
\end{equation}

\begin{figure}[t]
\centering
\includegraphics[scale=0.163,page=1]{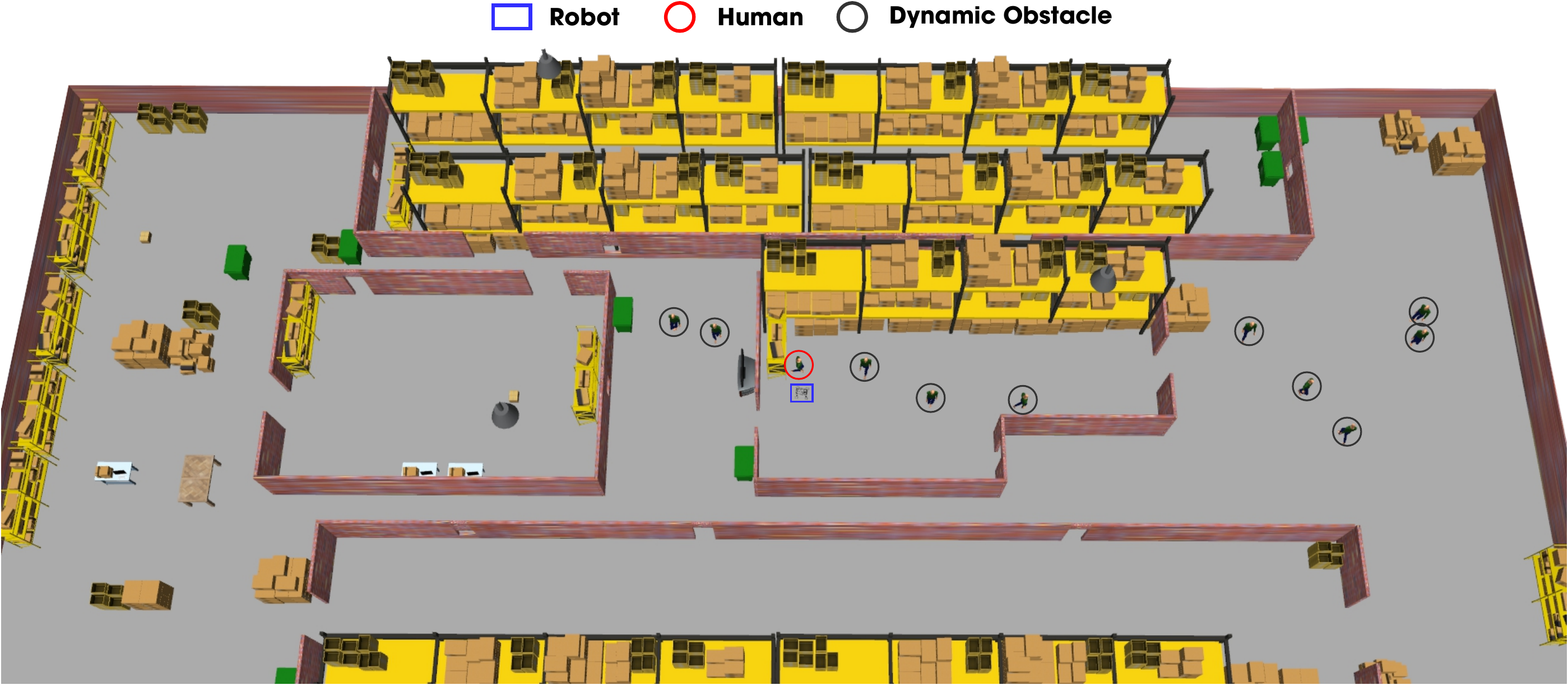}
\caption{Simulation environment used for comparison.}
\label{fig:gazebo}
\end{figure}
Due to the stochastic sampling in MPPI, a limited number of trajectory samples may yield control candidates that do not strictly satisfy the CBF condition. To address this issue, we introduce a local refinement step that filters the MPPI output $\mathbf{u}_k^+$. This refinement is formulated as a Quadratic Programming (QP) problem: 
\begin{align}
\mathbf{u}_k^{\text{safe}} = \arg\min_{\mathbf{u}} \quad & \frac{1}{2} \|\mathbf{u} - \mathbf{u}_k^+\|^2 \label{eq:local_repair} \\
\text{s.t.} \quad & \Delta h(\mathbf{q}^r_k, \mathbf{u}) \ge -\delta\, h(\mathbf{q}^r_k). \nonumber
\end{align}

The controller is implemented in Python using the Numba library \cite{lam2015numba} for GPU-accelerated parallel processing, with the parameters set as follows: 
$c_x=50.0$, $c_y=10.0$, $c_{\theta}=20.0$, $c_{\text{collision}}=1000.0$, $\lambda=1.0$, 
$\Delta t=0.1$, $K=100$, $M=2000$, $\delta=0.5$, and $c_{\text{safe}}=5000.0$. By leveraging parallel computation on the Jetson Orin, the system efficiently handles the computational workload. The MPPI controller achieves a stable average frequency of approximately 100 Hz, consuming about 5.82\% of CPU resources and 125.7 MB of GPU memory. Concurrently, the high-level RL policy operates at around 30 Hz and requires an additional computational footprint of 8.98\% CPU usage and 90.1 MB of GPU memory.
\begin{remark}
The robot treats the human as an obstacle through the MPPI obstacle avoidance cost function, which effectively prevents collisions but may still allow intrusion into the human intimate zone, causing discomfort and safety concerns. Therefore, a DCBF is introduced to enforce a minimum safety distance and prevent intrusion into the intimate zone.
\end{remark}
\section{Validation Results and Discussion}\label{sec:results}
In this section, the proposed method is validated through comparisons with other approaches and evaluated across multiple scenarios using the platform described in Section \ref{sec:system}.
\subsection{Comparison Results}
\begin{table}[t]
\centering
\caption{Quantitative Results of Comparison.}
\label{tab:compare}
\resizebox{\columnwidth}{!}{%
    \begin{tabular}{lccccc} 
    \toprule
    \midrule
    \textbf{Metrics} & \textbf{SR (\%)} & \textbf{UT (s)} & \textbf{$e_x$ (m)} & \textbf{$e_y$ (m)} & \textbf{$e_\theta$ (rad)} \\ 
    \midrule
    \textbf{PID \cite{chen2019human}} & 0 & - & - & - & - \\
    \midrule
    \textbf{DWA \cite{van2022collision}} & 34 & $33.0 \pm 8.5$ & $0.97 \pm 0.25$ & $0.88 \pm 0.22$ & $0.71 \pm 0.18$ \\
    \midrule
    \textbf{MPC \cite{peng2023mpc}} & 46 & $21.0 \pm 5.4$ & $0.41 \pm 0.13$ & $0.43 \pm 0.14$ & $0.23 \pm 0.06$ \\
    \midrule
    \textbf{DRL \cite{pang2020efficient}} & 62 & $27.0 \pm 6.3$ & $0.17 \pm 0.07$ & $0.21 \pm 0.11$ & $0.20 \pm 0.08$ \\
    \midrule
    \rowcolor{gray!20} \textbf{Proposed} & \textbf{86} & $\mathbf{2.5 \pm 0.8}$ & $\mathbf{0.09 \pm 0.03}$ & $\mathbf{0.05 \pm 0.02}$ & $\mathbf{0.08 \pm 0.03}$ \\
    \midrule
    \bottomrule
    \end{tabular}%
}
\end{table}
After successful training, we validated the proposed method, obtaining a tracking error of approximately 0.017 m for $|e_x|$, 0.021 m for $|e_y|$, and $e_\theta = 0.022$. To further evaluate the effectiveness of the proposed method, we compared it against PID \cite{chen2019human}, DWA \cite{van2022collision}, MPC \cite{peng2023mpc}, and RL \cite{pang2020efficient}. Experiments were conducted in the Gazebo simulation environment, as shown in Fig.~\ref{fig:gazebo}. The human moved through the workspace along trajectories covering both open and constrained areas, with speeds ranging from 0.7 m/s to 1.5 m/s. The environment included randomly placed static obstacles and 10 dynamic obstacles moving at varying speeds to increase complexity. Each method was evaluated over 50 trials using the following metrics: Success Rate (SR), defined as the proportion of successful trials (a failure occurs if the robot remains more than 2 m away from the human for over 5 s or if a collision occurs); Uncomfortable Time (UT), defined as the duration in which the robot enters the human’s intimate or social zones; and Tracking Error, computed as the RMSE of position and orientation deviations $(e_x, e_y, e_\theta)$. It is also notable that all compared methods are evaluated under a mapless setting, where no prior map of the environment is provided and no knowledge of the future target trajectory is assumed. Furthermore, for a fair comparison, all methods operate solely at the velocity-command level, without the use of behavior trees, global planning modules, or any additional high-level planning strategies.

Table~\ref{tab:compare} presents the quantitative results, including the mean and standard deviation (SD). The results show that the proposed method achieves at least a 24\% improvement in success rate and a 47\% improvement in tracking accuracy, while also reducing discomfort time, demonstrating its ability to adapt tracking positions to environmental constraints and provide flexible, socially compliant accompaniment. The PID controller fails to complete the task due to its lack of obstacle-avoidance capability. Although DWA and MPC incorporate obstacle avoidance, their reliance on fixed tracking positions leads to larger errors and lower success rates in constrained environments, resulting in target loss and increased discomfort. The RL baseline improves obstacle avoidance but still yields higher discomfort time, as it tends to prioritize safety margins over social norms and may intrude into the user's personal space.
\begin{figure*}[t]
\centering
\includegraphics[scale=0.335,page=1]{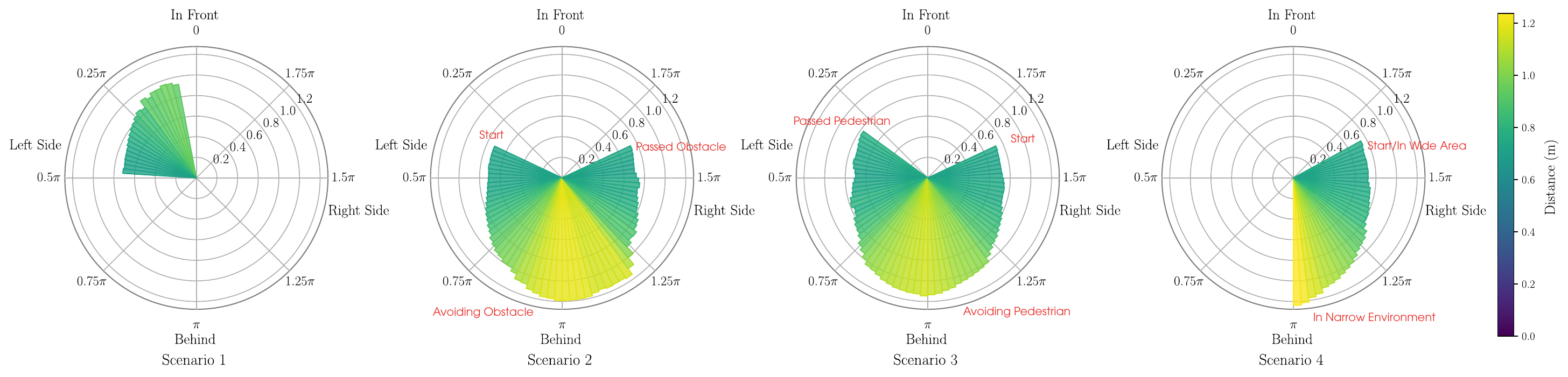}
\caption{Distribution of robot positions relative to the human across four experimental scenarios.}
\label{fig:polar}
\end{figure*}
\begin{figure}[t]
\centering
\includegraphics[scale=0.42,page=1]{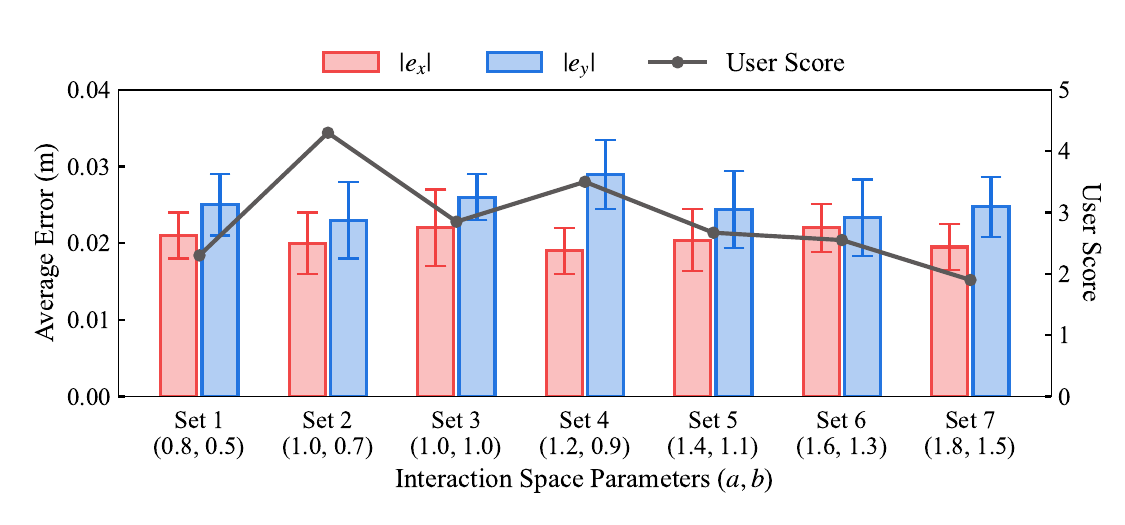}
\caption{Tracking performance and user scores under different interaction-space configurations.}
\label{fig:interactive_space}
\end{figure}

Furthermore, in terms of computational efficiency, although the PID controller exhibits the lowest computational cost with an operating frequency of approximately $200\,\text{Hz}$, it completely fails to complete the task. Compared with tracking strategies such as DWA ($\approx 41\,\text{Hz}$), MPC ($\approx 35\,\text{Hz}$), and DRL ($\approx 27\,\text{Hz}$), our high-level planning strategy maintains an operating frequency of approximately $30\,\text{Hz}$. Although slightly lower than DWA and MPC, this frequency is fully sufficient for updating adaptive companion positions, as the module is responsible only for spatial decision-making rather than continuous motor control. The low-level MPPI controller achieves an operating frequency of $100\,\text{Hz}$, which is significantly faster than the other methods.

Despite achieving the best performance, the proposed method still exhibits a 14\% failure rate. In some cases, the robot becomes temporarily trapped among obstacles. Since the controller prioritizes safety and maintaining a socially acceptable distance from the human, it may stop or take a longer detour. If the robot remains more than 2 meters away from the human for over 5 seconds, the episode is counted as a failure due to loss of tracking. These results highlight the need for a more adaptive tracking mechanism, such as incorporating the relative velocity and motion direction of dynamic obstacles into the RL state to enable more proactive decision-making.
\subsection{Interaction Space Evaluation Experiments}
To evaluate the adaptability of the proposed method under varying interaction spaces, experiments were conducted across seven interaction-space configurations with 10 participants aged 20–30 walking at a normal speed. It is worth noting that no fine-tuning was performed during sim-to-real transfer. A user survey was also conducted to identify the most preferred configuration. The results are shown in Fig.~\ref{fig:interactive_space}. The results show that, despite significant variations in space dimensions, the tracking errors remained stable, with $|e_x|$ and $|e_y|$ ranging from approximately $0.019$ m to $0.022$ m and $0.023$ m to $0.029$ m, respectively. This demonstrates the strong generalization capability of the DRL policy with domain randomization, enabling adaptation to different spatial constraints while maintaining accurate control. In addition, circular interaction spaces (e.g., Set 3) received lower user preference scores than elliptical ones (e.g., Set 2), consistent with human locomotion characteristics that require a larger frontal safety margin. Moreover, configurations within the Personal Zone (Sets 1–4) achieved higher satisfaction than those within the Social Zone (Sets 5–7). Users reported a trade-off in perception, where larger distances increased concerns about losing track of the user, whereas smaller configurations (e.g., Set 1) raised concerns about potential collisions due to limited spacing. Set 2 achieved the highest user preference score and was selected for subsequent experiments.
\subsection{Indoor Experiments}
\subsubsection{Tracking Evaluation}
The effectiveness of the proposed method was evaluated through four indoor scenarios, conducted without predefined tracking points and with the robot’s initial position set freely. Scenario 1 evaluated baseline tracking in an open environment without obstacles, where the robot followed a human along a rectangular path with sharp turns. Scenario 2 introduced a static obstacle on the left, constraining the workspace. Scenario 3 involved dynamic interaction, where a pedestrian approached from the front-right while the robot followed the human. Scenario 4 further reduced navigable space as both lateral boundaries gradually narrowed, creating a confined environment.

\begin{figure*}[t]
\centering
\includegraphics[scale=0.54,page=1]{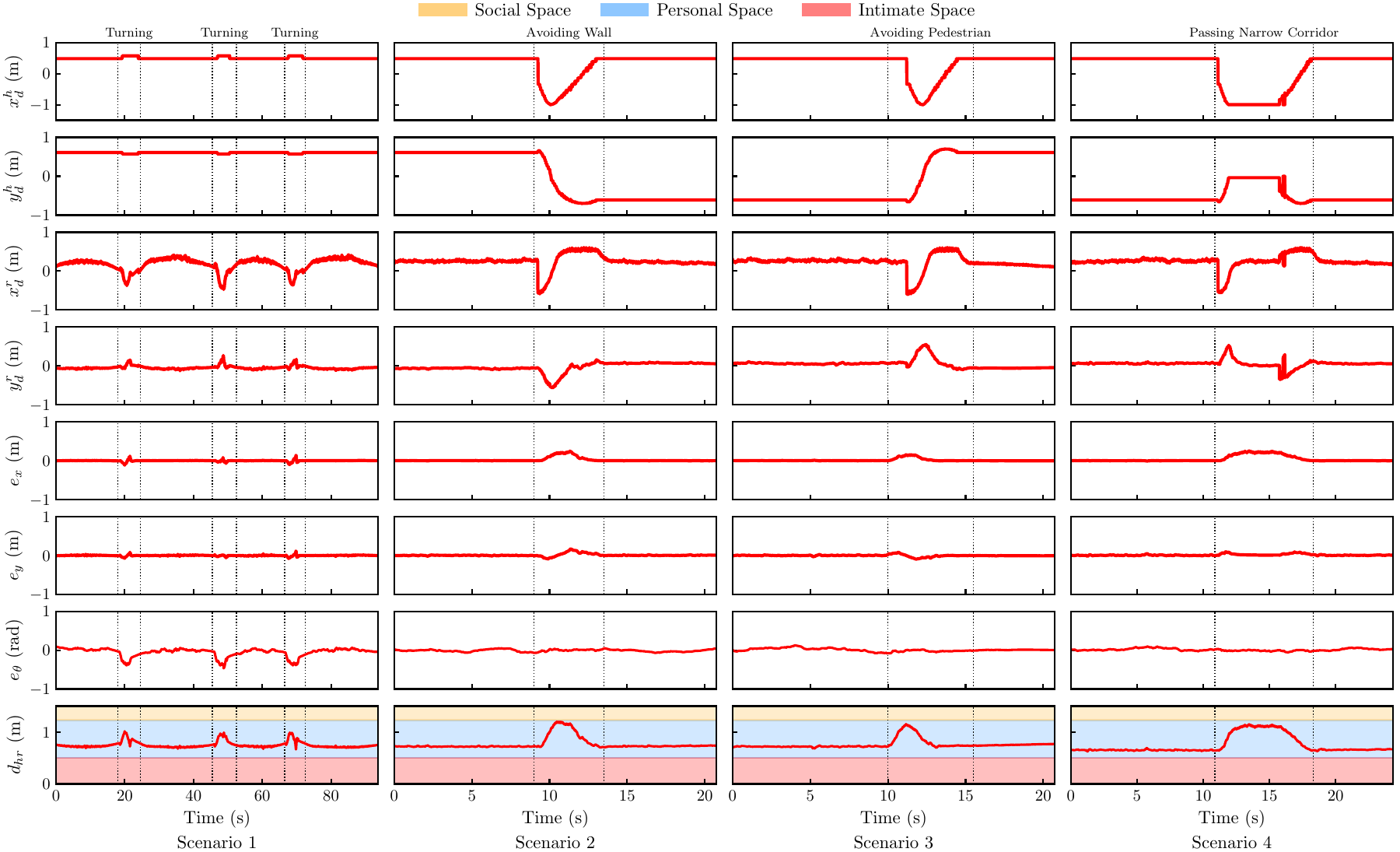}
\caption{Tracking performance of the proposed controller in indoor scenarios: relative positions of the following point to the human ($x_d^h$, $y_d^h$) and the robot ($x_d^r$, $y_d^r$), tracking errors ($e_x$, $e_y$, $e_\theta$), and the distance between the human and the robot ($d_{hr}$).}
\label{fig:line}
\end{figure*}
Fig.~\ref{fig:polar} depicts the spatial distribution of the robot relative to the human across four experimental scenarios. In Scenario 1, the robot primarily stayed on the left side but occasionally moved toward the front, indicating that the proposed method flexibly adapts to human motion and environmental changes, allowing smooth transitions between side-by-side and frontal accompaniment. In Scenarios 2 and 3, the robot switched between the left and right sides in opposite directions, showing that it autonomously adjusted its lateral position to maintain continuous following without getting trapped, even in constrained environments with static or moving obstacles.

Notably, in Scenarios 2 and 3 the robot did not return to its original side but instead shifted laterally to avoid obstacles, demonstrating effective adaptability in tight spaces. In contrast, in Scenario 4 the robot resumed the right-side position rather than switching sides, highlighting its ability to make context-dependent decisions based on real-time environmental conditions rather than relying on fixed behavioral patterns.

This adaptability was further validated through the analysis of tracking performance, as illustrated in Fig.~\ref{fig:line}. The tracking points relative to the robot $(x_d^r, y_d^r)$ continuously changed in response to workspace constraints and human actions, rather than maintaining a fixed position as in previous studies. These points evolve smoothly instead of exhibiting abrupt oscillations, indicating that the policy enables gradual and stable motion adaptation based on observations. In addition, the tracking errors ($e_x, e_y, e_\theta$) converge toward zero even when the robot must adjust its strategy to avoid obstacles, demonstrating the effectiveness of integrating the tracking controller with the policy. The position error along the $x$-axis in Scenario 4 was larger than in the other three scenarios, indicating that when the robot followed from behind, the deviation in the $x$-direction increased, whereas in the other three scenarios, the error fluctuated around zero. Additionally, the angular error in Scenario 1 was larger than in the other scenarios due to the sudden changes in the human’s movement direction. Meanwhile, the position errors along the $y$-axis in all four scenarios remained similar, oscillating around zero. 

\begin{figure*}[t]
\centering
\includegraphics[scale=0.35,page=1]{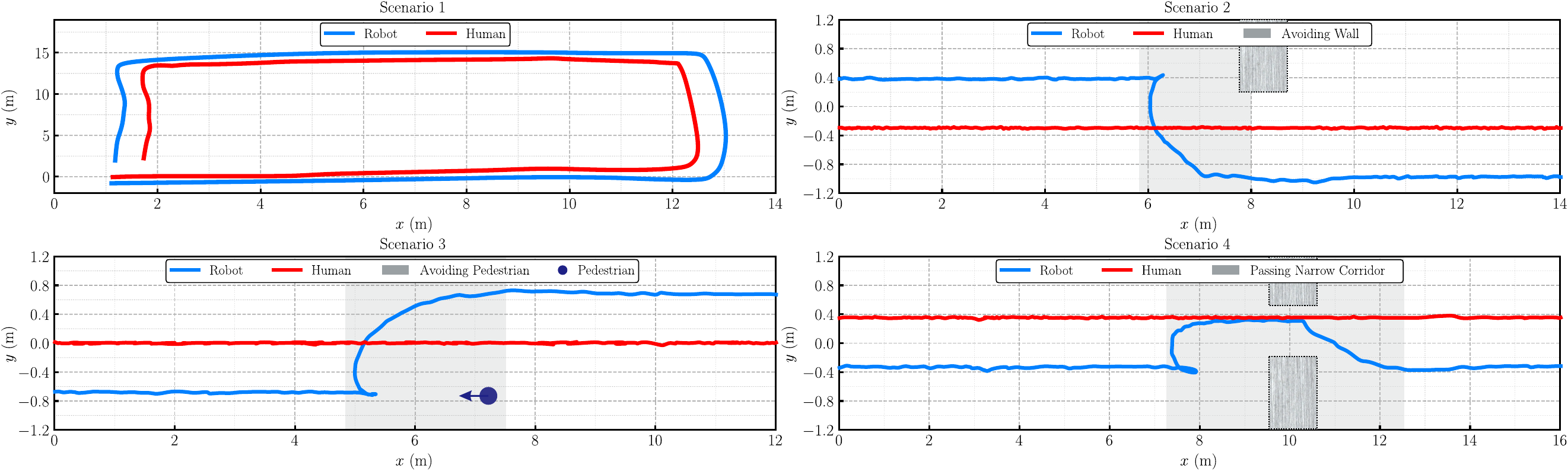}
\caption{Trajectories of the human and robot in four indoor scenarios.}
\label{fig:trajectory}
\end{figure*}

It is also notable that the tracking errors ($e_x, e_y, e_\theta$) tended to increase during positional or angular transitions. While these deviations may have been momentarily large, they gradually converged to zero as the robot stabilized its tracking behavior. Furthermore, it is noteworthy that across all four scenarios, the robot’s position distribution relative to the human aligned with the designed elliptical interactive space (Fig.~\ref{fig:polar}) rather than following a random pattern. This adherence indicated compliance with social Proxemics principles, as the human-robot distance ($d_{hr}$) remained within the personal space range of 0.7–1.15 m. In Scenario 1, despite significant directional changes, the robot maintained this spacing without intruding into the intimate zone, ensuring safety and adherence to social norms. Similarly, in other scenarios, even when adjusting its tracking position, the robot remained within the designated personal space without exceeding its limits. These motion patterns are also characterized by the movement trajectories, as shown in Fig.~\ref{fig:trajectory}.
\begin{figure*}[t]
\centering
\includegraphics[scale=0.35,page=1]{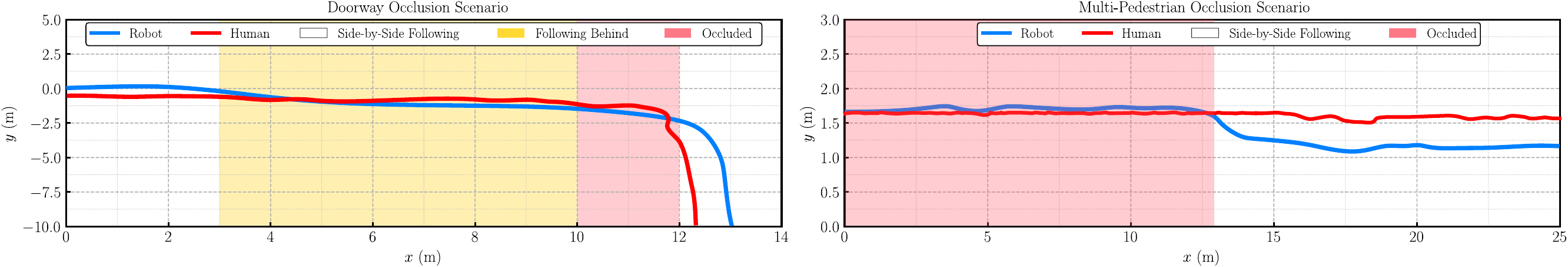}
\caption{Human and robot trajectories in indoor occlusion scenarios (doorway and multi-pedestrian).}
\label{fig:trajectories_occlusion}
\end{figure*}
\subsubsection{Occlusion Experiments}
To further evaluate system robustness, we conducted two indoor experiments under occlusion in typical human–robot accompaniment scenarios. The doorway occlusion scenario tested tracking performance when the human passed through a doorway and made a sudden turn, temporarily blocking the robot’s view. The multi-pedestrian occlusion scenario evaluated whether the robot could correctly maintain and recover the identity of the sensor-equipped user when three pedestrians occluded the target.

Fig.~\ref{fig:trajectories_occlusion} shows the trajectories for both scenarios. In the doorway case, the robot initially followed side by side, then moved behind the human when entering a narrow corridor. Despite temporary loss of visibility at the doorway, the robot maintained an accurate pose estimate, passed through the doorway without collision, and returned to a side-by-side formation once space allowed. In the multi-pedestrian case, the robot maintained safe distance and correct target identification throughout the occlusion. After the pedestrians cleared, it accelerated to regain a side-following position. These results demonstrate that the proposed method achieves robust human-following under significant occlusions.

To further validate the robustness of the proposed control framework, additional comparative experiments were conducted against Camera- and LiDAR-based tracking systems under occlusion conditions. The results in Table~\ref{tab:tracking_compare} show clear performance differences in cluttered and highly dynamic environments. Among all evaluated methods, AmfiTrack achieves the best performance, with a Success Rate of 92\%, enabled by its electromagnetic sensing-based localization mechanism, which supports continuous re-identification and maintenance of the user coordinate frame even under full occlusion caused by static structures (e.g., walls) or dynamic obstacles (e.g., crossing pedestrians). In contrast, LiDAR achieves only 39\% Success Rate due to its reliance on geometric point-cloud representations, which provide limited discriminative information and lead to false positives and target confusion. The Camera-based method reaches 55\%, being constrained by optical limitations, where visual ambiguity between the target user and surrounding pedestrians frequently results in tracking loss and ID switching.
\begin{table}[t]
\centering
\caption{Quantitative Performance Comparison of Tracking Hardware Systems.}
\label{tab:tracking_compare}
\resizebox{\columnwidth}{!}{%
    \begin{tabular}{lcccc}
    \toprule
    \midrule
    \textbf{Method} & \textbf{Success Rate (\%)} & \textbf{Frequency (Hz)} & \textbf{Lighting Robustness} & \textbf{FoV} \\
    \midrule
    \textbf{Camera} & 55 & $\approx 30$ & Low & $110^\circ$ \\
    \midrule
    \textbf{LiDAR} & 39 & $\approx 9$ & High & $360^\circ$ \\
    \midrule
    \rowcolor{gray!20} \textbf{AmfiTrack} & \textbf{92} & $\approx \mathbf{120}$ & \textbf{High} & \textbf{Omni} \\
    \midrule
    \bottomrule
    \end{tabular}%
}
\end{table}

In terms of real-time performance, AmfiTrack operates at approximately 120 Hz, which is 4$\times$ higher than Camera (~30 Hz) and more than 10$\times$ higher than LiDAR (~9 Hz). This substantially reduces control-loop latency and improves trajectory smoothness and responsiveness in close-range human–robot interaction. In contrast, both Camera and LiDAR suffer from heavy processing pipelines and strong dependence on feature extraction, leading to increased latency and degraded performance in dynamic environments. Regarding robustness to illumination changes, Camera is highly sensitive to lighting variations, whereas LiDAR and AmfiTrack remain more stable due to independence from optical intensity. In terms of FoV, LiDAR provides full 360$^\circ$ coverage, Camera is limited to approximately 110$^\circ$, while AmfiTrack achieves omnidirectional (omni) sensing. Overall, these properties collectively enable AmfiTrack to consistently outperform both baseline systems across all evaluated scenarios.
\begin{figure*}[t]
\centering
\includegraphics[scale=0.163,page=1]{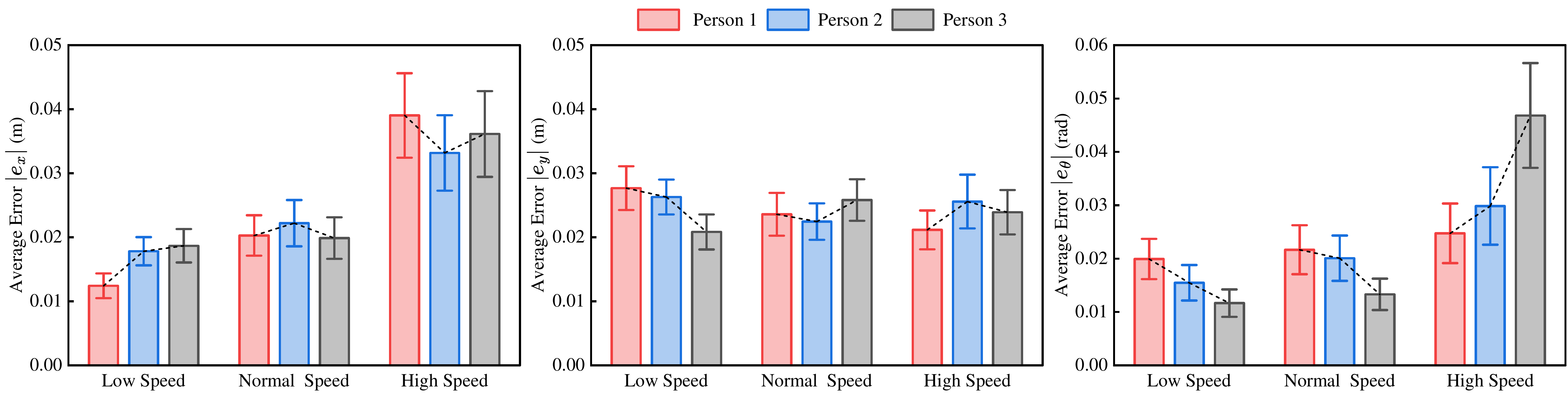}
\caption{Tracking accuracy at different walking speeds.}
\label{fig:vel_3p}
\end{figure*}
\begin{figure*}[t]
\centering
\includegraphics[scale=0.143,page=1]{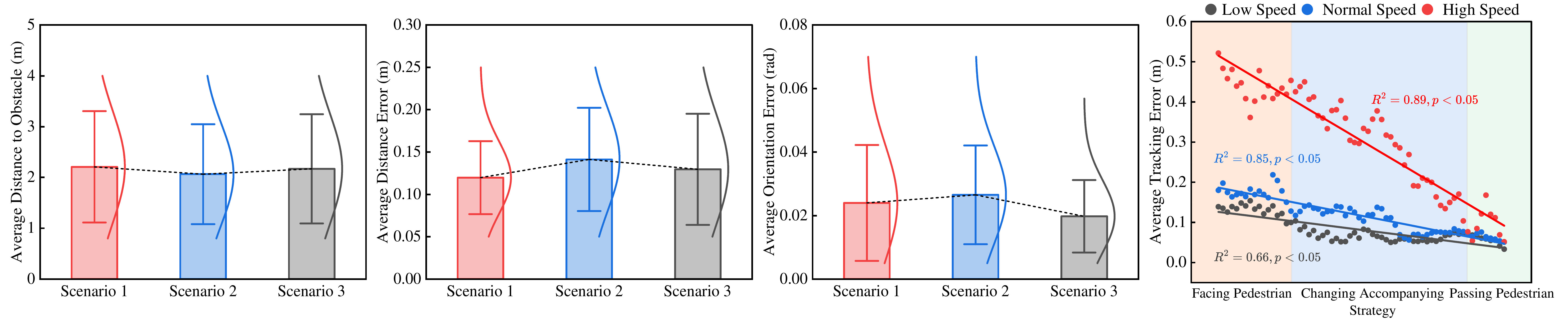}
\caption{Tracking performance under various obstacle conditions.}
\label{fig:box_linear}
\end{figure*}
\begin{figure}[t]
\centering
\includegraphics[scale=0.235,page=1]{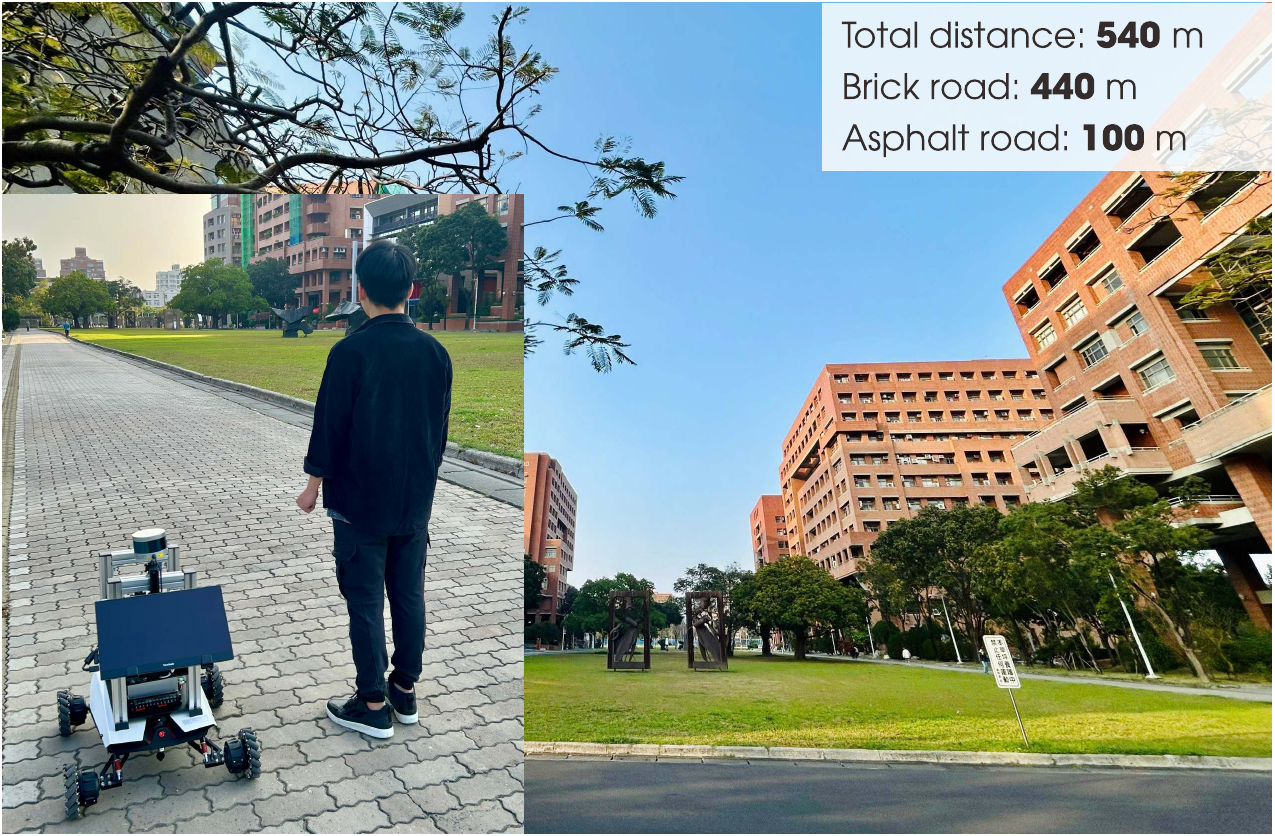}
\caption{Outdoor experimental environments.}
\label{fig:outdoor_exp}
\end{figure}
\subsection{Outdoor Experiments}
To explore future applications of human-accompanying tasks for mobile robots, we conducted validation experiments in an outdoor urban environment, involving pedestrians and running individuals around the robot, which act as dynamic obstacles, as well as a workspace with varying configurations. In particular, we further evaluated the system under two different time periods, namely daytime and nighttime, in order to assess its performance under varying lighting conditions. The experiments were carried out on everyday terrains, including straight paths and turns, with two common types of surfaces: asphalt and brick roads. The experimental site covered a total distance of 540 m (Fig.~\ref{fig:outdoor_exp}), comprising 440 m of brick road and 100 m of asphalt road. Three experiments were conducted: the first experiment evaluated the operational speed range of the proposed system. Three participants participated in this study, each walking the entire distance at three distinct speeds. The speed levels were defined as low speed (average 0.75 m/s), normal speed (average 1.2 m/s, representing typical walking speed), and high speed (average 1.7 m/s). The second experiment evaluated the robot’s performance on brick and asphalt roads while the human maintained a normal walking speed. The third experiment examined how speed affected the robot’s ability to adjust its tracking position for obstacle avoidance. Initially, three obstacle scenarios were tested at normal speed, followed by additional trials in which walking speed was varied within a single scenario to analyze its impact on tracking performance.

The average tracking errors of $\left| e_x \right|$, $\left| e_y \right|$, and $\left| e_\theta \right|$ for the three participants while walking at three different experimental speeds are presented in Fig.~\ref{fig:vel_3p}. The mean values of $\left| e_x \right|$ and $\left| e_y \right|$ across all three speed ranges remained below 4 cm and 3 cm, respectively, while the angular error $\left| e_\theta \right|$ was less than 0.06 rad. Additionally, the error variations among the three participants across different speeds were consistent. Notably, positional errors were more pronounced along the $x$-axis than along the $y$-axis at the high speed. Although the third participant exhibited slightly higher angular errors at high speed than the other two, the difference was not significant.

\begin{figure}[t]
\centering
\includegraphics[scale=0.305,page=1]{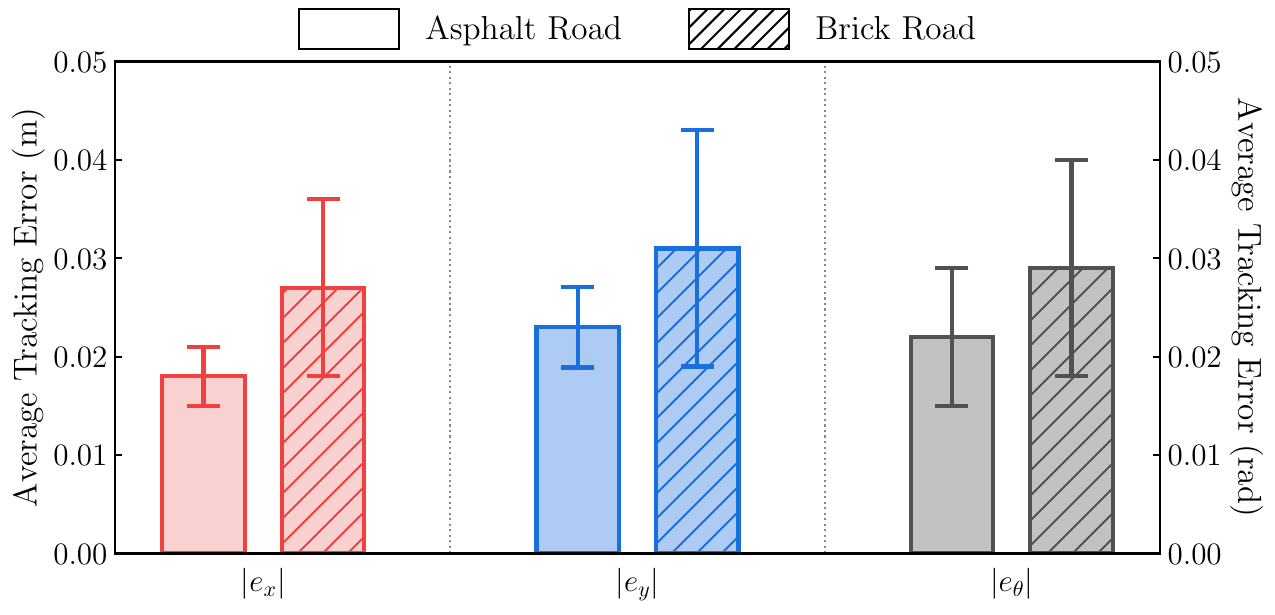}
\caption{Tracking performance on asphalt and brick roads.}
\label{fig:ground}
\end{figure}

The tracking performance in terms of $e_x$, $e_y$, and $e_\theta$ on both asphalt and brick roads is illustrated in Fig.~\ref{fig:ground}. The results show that tracking on asphalt surfaces exhibits stability compared to brick roads, particularly in $e_x$ and $e_\theta$, although the differences are not significant. This indicates that both surface types maintain acceptable error levels during movement.

Fig.~\ref{fig:box_linear} illustrates the performance of the proposed method under obstacle conditions. The robot maintained average distances of $2.21 \pm 1.09$ m and $2.17 \pm 1.07$ m from obstacles in the first and third scenarios, respectively, while the second scenario showed a slightly lower average distance of $2.06 \pm 0.98$ m. Consequently, the second scenario exhibited a slightly higher average distance error compared to the others, with a mean error of $0.14 \pm 0.06$ m and an average angular error of $0.026 \pm 0.015$ rad. A linear regression analysis was conducted to examine the relationship between walking speed and tracking adjustment for obstacle avoidance. The results indicate a statistically significant correlation between obstacle distance and tracking error across three speed ranges, with $R^2 = 0.66$, $R^2 = 0.85$, and $R^2 = 0.89$, all with $p < 0.05$.
\subsection{Crowded Environment Experiments}
To evaluate the expandability of the proposed method in crowded environments, we conducted experiments by increasing the number of dynamic pedestrians from 5 to 20, as summarized in Table~\ref{tab:crowd_density}. The quantitative assessment was performed using three metrics: SR, UT, and Average Planning Time (APT). The results show that as crowd density increases, SR gradually decreases, reaching 81\% with 5 pedestrians, 65\% with 10 pedestrians, and then dropping sharply to 33\% and 26\% with 15 and 20 pedestrians, respectively, while UT increases significantly. This degradation occurs because, in a crowded environment, the robot must prioritize obstacle avoidance, which forces frequent trajectory adjustments and pushes the robot farther away from the user. These effects increase the likelihood of target loss or collisions. This noticeable decline in performance highlights the need for further research on companion strategies that can maintain continuous and safe tracking in extremely dense pedestrian environments. Despite this, APT remains remarkably stable as the number of pedestrians increases, fluctuating only within a narrow range from $9.86\text{ ms}$ to $11.85\text{ ms}$.
\begin{table}[t]
\centering
\caption{Quantitative Performance across Crowd Densities.}
\label{tab:crowd_density}
\renewcommand{\arraystretch}{1.2} 
\begin{tabular}{||>{\centering\arraybackslash}m{1.75cm}||>{\centering\arraybackslash}m{1.45cm}||>{\centering\arraybackslash}m{1.45cm}||>{\centering\arraybackslash}m{1.45cm}||} 
\hline
\textbf{Pedestrians} & \textbf{SR (\%)} & \textbf{UT (s)} & \textbf{APT (ms)} \\ 
\hline\hline
\textbf{5} & 81 & 1.35 & 9.86 \\
\hline
\textbf{10} & 65 & 2.79 & 10.01 \\
\hline
\textbf{15} & 33 & 7.13 & 10.93 \\
\hline
\textbf{20} & 26 & 9.22 & 11.85 \\
\hline
\end{tabular}
\end{table}
\subsection{Practical Considerations and Future Directions}
Despite the proposed method having been tested in both indoor and outdoor scenarios, the current study acknowledges several limitations that will be addressed in future work. The dimensions of the interaction space are primarily determined based on a user survey involving 10 participants and the selection of the most preferred interpersonal distance. Future work should focus on enabling the interaction space to be automatically adapted to individuals with different body sizes, ages, personalities, walking speeds, genders, and cultural backgrounds, as well as increasing the number of participants in the experimental evaluation. Although the proposed method has been tested in real-world scenarios and evaluated in environments with multiple people, navigation in densely crowded environments with high pedestrian density remains a significant challenge that requires further investigation. Moreover, a slight increase in tracking error was observed on brick surfaces, indicating a degradation in the performance of the mecanum wheels on uneven terrain. This suggests a potential area for hardware improvement.
\section{Conclusion}\label{sec:conclusion}
In this study, we proposed a novel human-companion robot framework for navigation in diverse environments without relying on fixed tracking positions. The proposed method adaptively switches among front, side, and rear tracking positions according to the surrounding context. A hierarchical control architecture combines a PPO-based RL policy for socially appropriate tracking-position selection with an MPPI-CBF controller for safe and smooth motion execution. Experimental results demonstrate lower tracking errors and higher success rates than fixed-position baselines, while real-world experiments validate the framework in both indoor and outdoor environments. Despite these promising results, several limitations remain. The interaction-space parameters are manually defined and should be personalized according to user characteristics, such as age, personality, walking speed, gender, and cultural background. In addition, the experiments involve a limited number of participants, and performance degrades in highly crowded environments and on uneven terrain due to the limitations of the mecanum-wheel platform. Future work will focus on learning-based interaction-space personalization, improving robustness in crowded environments, and conducting larger-scale user studies.
\bibliographystyle{IEEEtran}
\bibliography{reference}
\begin{IEEEbiography}[{\includegraphics[width=1in,height=1.25in,clip,keepaspectratio]{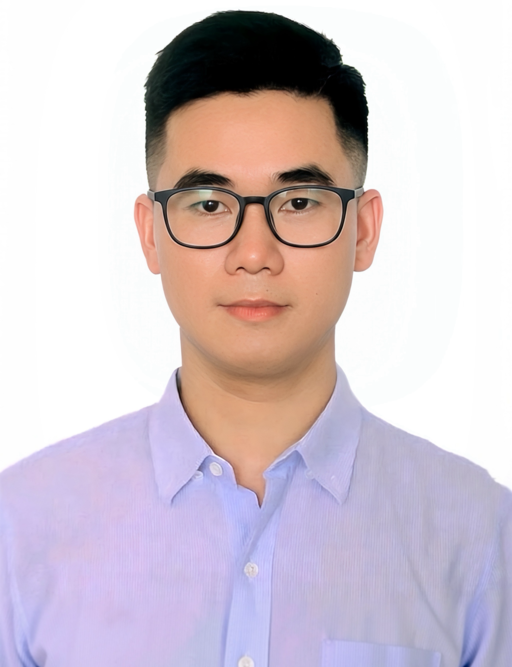}}]{Cong-Thanh Vu} (Student Member, IEEE) received the B.S. and M.S. degrees in Mechatronics Engineering from Hanoi University of Industry, Hanoi, Vietnam, in 2019 and 2021, respectively. He is currently pursuing a Ph.D. at the Networked Robotic Systems Laboratory (NRSL), Department of Mechanical Engineering, National Cheng Kung University (NCKU), Tainan, Taiwan.

His research interests focus on socially assistive robotics, human–robot interaction, robot learning, and agricultural robotics.
\end{IEEEbiography}
\begin{IEEEbiography}[{\includegraphics[width=1in,height=1.25in,clip,keepaspectratio]{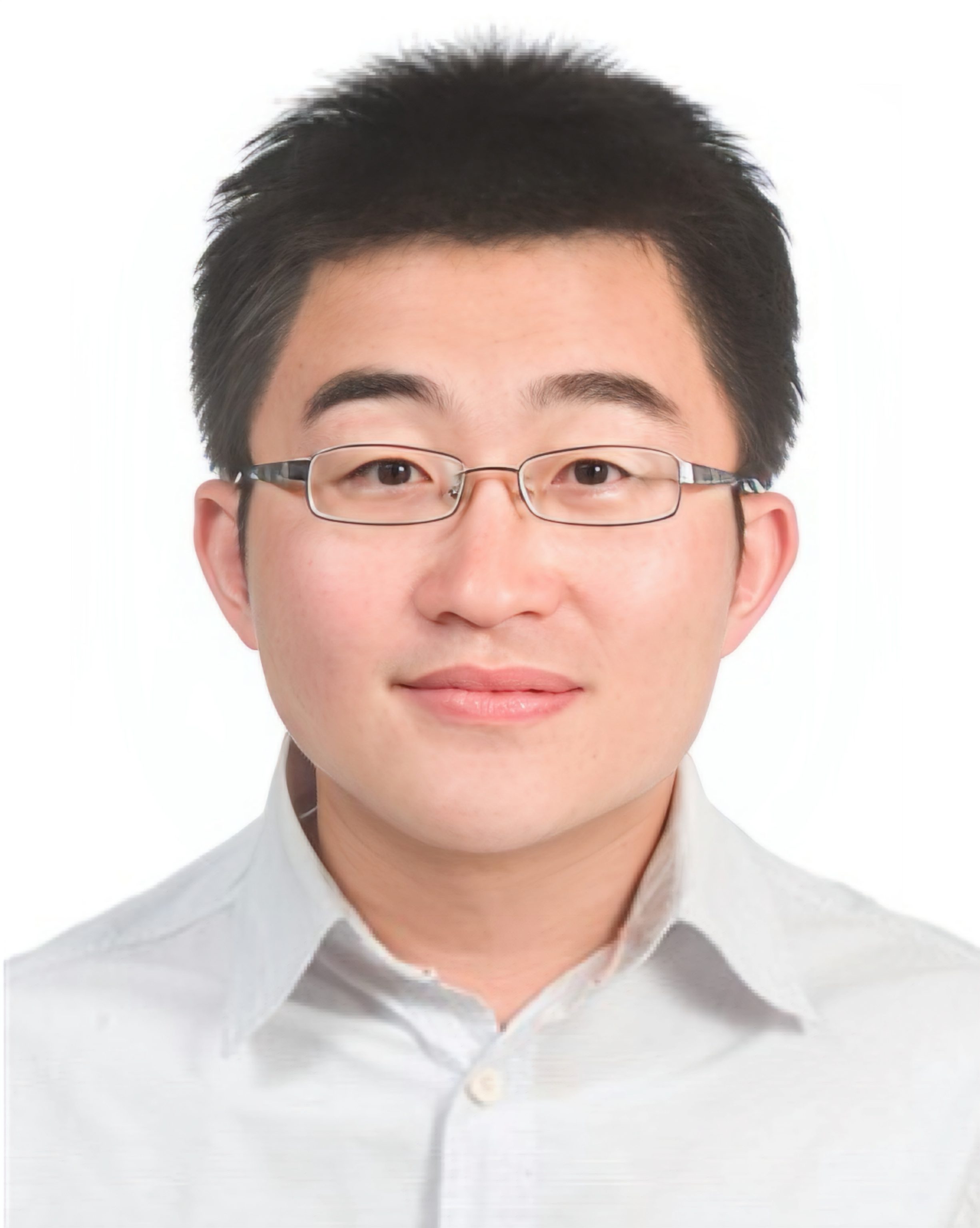}}]{Yen-Chen Liu} (Senior Member, IEEE) received the B.S. and M.S. degrees in mechanical engineering from National Chiao Tung University, Hsinchu, Taiwan, in 2003 and 2005, respectively, and the Ph.D. degree in mechanical engineering from the University of Maryland, College Park, MD, USA, in 2012.

He is currently a Distinguished Professor with the Department of Mechanical Engineering, National Cheng Kung University, Tainan, Taiwan. His research interests include the control of networked robotic systems, multiagent systems, mobile robot networks, human–robot interaction, and artificial intelligence (AI)-driven mobile robots.

Dr. Liu was a recipient of the Ta-You Wu Memorial Award, Ministry of Science and Technology (MOST), Taiwan, in 2016, the Kwoh-Ting Li Research Award, National Cheng Kung University, Taiwan, in 2018, the Young Scholar Fellowship-Columbus Program, MOST, Taiwan, in 2019, and Outstanding Research Award, National Science and Technology Council (NSTC), Taiwan, in 2023.
\end{IEEEbiography}
\end{document}